\documentclass[10pt,twocolumn,letterpaper]{article}

\usepackage[pagenumbers]{wacv} % To force page numbers, e.g. for an arXiv version

\usepackage{graphicx}
\usepackage{amsmath}
\usepackage{amssymb}
\usepackage{booktabs}
\usepackage{amsfonts}
\usepackage{bm}
\usepackage{enumerate}
\usepackage{tabularx}
\usepackage{gensymb}
\usepackage{multirow}

\usepackage{orcidlink}
\usepackage{color,soul}
\usepackage{xcolor}
\definecolor{cerulean}{rgb}{0.0,0.48,0.65}
\definecolor{green}{rgb}{0.01, 0.75, 0.24}
\definecolor{Black}{RGB}{0.0, 0.0, 0.0}
\newcommand{\s}[1]{\textcolor{red}{#1}}

\newcommand{\blue}[1]{\textcolor{blue}{#1}}

\newcommand{\shadow}[1]{}
\def\s{\s}
\def\b{\blue}

\def\s{\shadow}

\usepackage[capitalize]{cleveref}
\crefname{section}{Sec.}{Secs.}
\Crefname{section}{Section}{Sections}
\Crefname{table}{Table}{Tables}
\crefname{table}{Tab.}{Tabs.}

\usepackage{tikz}
\def\wacvPaperID{2176} % *** Enter the WACV Paper ID here
\def\confName{WACV}
\def\confYear{2025}

\begin{document}

\title{Cross-Task Affinity Learning for Multitask Dense Scene Predictions}

\author{Dimitrios Sinodinos$^{1,2}$, Narges Armanfard$^{1,2}$\\
$^{1}$McGill University, $^{2}$Mila - Quebec AI Institute\\
Montreal, Canada\\
{\tt\small dimitrios.sinodinos@mail.mcgill.ca, narges.armanfard@mcgill.ca}
% \and
% {\tt\small narges.armanfard@mcgill.ca}
}
\maketitle

% Insert a text box in the bottom-left corner of the first page text area
\begin{tikzpicture}[remember picture, overlay]
    \node[anchor=south west, xshift=0.75in, yshift=0.25in] at (current page.south west) {%
        \footnotesize Accepted for publication at WACV25
    };
\end{tikzpicture}

%%%%%%%%% ABSTRACT
\begin{abstract}
\s{Multitask learning (MTL) has gained prominence for its ability to jointly predict multiple tasks, achieving better per-task performance while using fewer per-task model parameters than single-task learning. More recently, decoder-focused architectures have considerably improved multitask performance by refining task predictions using the features of other related tasks. However, most of these refinement methods fail to simultaneously capture local and long-range dependencies between task-specific representations, as well as cross-task patterns in a parameter-efficient manner. In this paper, we introduce the Cross-Task Affinity Learning module, which uses a lightweight framework that enhances the task refinement capabilities of multitask networks. CTAL adeptly captures local and long-range cross-task interactions by manipulating task affinity matrices in a manner that is optimally suited to apply parameter-efficient grouped convolutions without worrying about information loss. Our results show that we achieve state-of-the-art MTL performance among decoder-focused algorithms for both CNN and transformer backbones, while using substantially fewer model parameters than single-task learning.}
Multitask learning (MTL) has become prominent for its ability to predict multiple tasks jointly, achieving better per-task performance with fewer parameters than single-task learning. Recently, decoder-focused architectures have significantly improved multitask performance by refining task predictions using features from related tasks. However, most refinement methods struggle to efficiently capture both local and long-range dependencies between task-specific representations and cross-task patterns. In this paper, we introduce the Cross-Task Affinity Learning (CTAL) module, a lightweight framework that enhances task refinement in multitask networks. CTAL effectively captures local and long-range cross-task interactions by optimizing task affinity matrices for parameter-efficient grouped convolutions without concern for information loss. Our results demonstrate state-of-the-art MTL performance for both CNN and transformer backbones, using significantly fewer parameters than single-task learning.
\end{abstract}

%%%%%%%%% BODY TEXT
\section{Introduction}
\label{sec:intro}

AI research is rapidly advancing, but many cutting-edge models are too large for deployment on edge devices like phones or wearables, which rely on remote access. For many applications, local operation without network dependence underscores the need to balance performance with model efficiency. In recent years, multitask learning (MTL)~\cite{caruana1997multitask} has gained attention as a parameter-efficient paradigm that often outperforms single-task learning (STL). MTL typically involves a single network optimized for multiple tasks by jointly minimizing several loss functions. This leads to shared layers or features across tasks.
\s{AI research is rapidly integrating into daily life. However, most cutting-edge models are massive and rely on remote access since they are difficult to deploy on edge devices, such as mobile phones, smart accessories, or wearable medical equipment. For many applications, the need for models to operate locally without network dependence highlights the importance of balancing performance with parameter efficiency in model design.}
%
% Introduce the idea of refining task predictions in multitask networks
\s{
In recent years, multitask learning (MTL)~\cite{caruana1997multitask} has emerged as a parameter-efficient learning paradigm that can also outperform traditional single-task learning (STL). Generally, MTL involves a single network that can learn multiple tasks by jointly optimizing multiple loss functions. Consequently, using a single network means that several layers or features are being shared between tasks.} 
\s{Unlike tasks like image classification, where a single label is assigned to an entire image, dense prediction involves assigning labels or values to every pixel.
MTL is useful in dense prediction tasks because it enables a model to learn shared representations across related tasks, improving performance and efficiency. In dense prediction, tasks like semantic segmentation, depth estimation, and surface normals estimation often share common low-level features (e.g., edges, textures). By learning these tasks together, MTL allows the model to generalize better by leveraging shared information, reducing overfitting, and improving robustness. Additionally, MTL can reduce the need for task-specific models, making the system more computationally efficient.}\s{In many dense prediction cases, sharing features across tasks has been proven to improve per-task performance, while using fewer per-task model parameters. This is the result of improved generalization by leveraging domain-specific knowledge between related tasks~\cite{kendall2018multi}.}

Unlike image classification, where a single label applies to an entire image, dense prediction assigns labels to every pixel. MTL benefits dense prediction tasks by enabling shared representations across related tasks, enhancing both performance and efficiency. Tasks like semantic segmentation, depth estimation, and surface normals estimation share low-level features (e.g., edges, textures), and learning them together helps the model generalize better, reduce overfitting, and improve robustness. Additionally, MTL reduces the need for separate task-specific models, increasing computational efficiency.

\begin{figure*}[htb!]
    \centering
    \includegraphics[width=\linewidth]{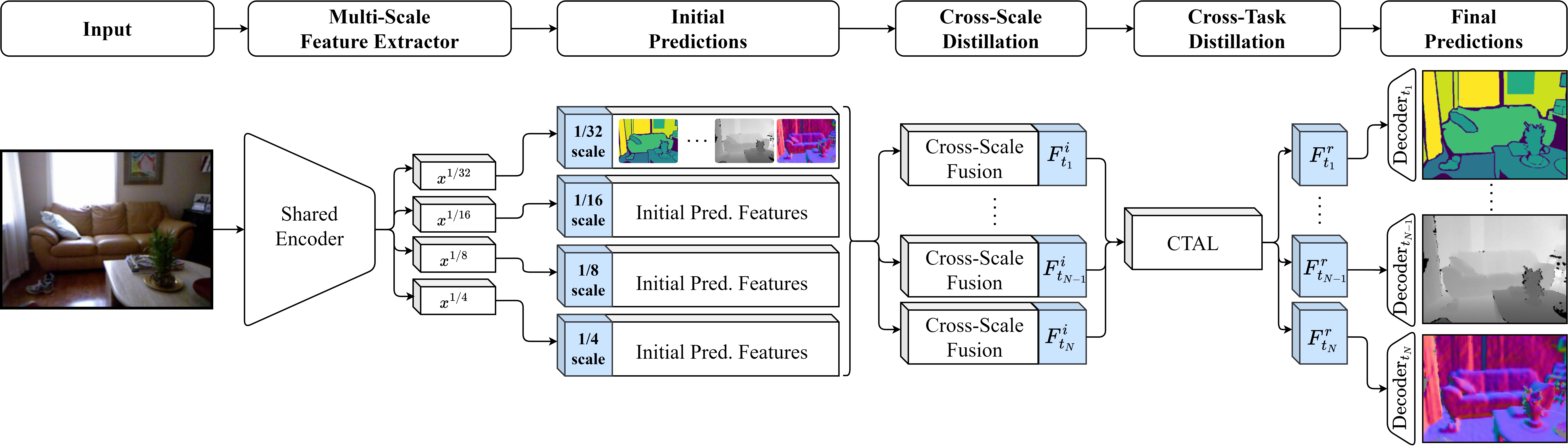}
    \caption{A network diagram of the task-prediction distillation framework using deep supervision at multiple feature scales and using the CTAL module after cross-scale fusion for task-refinement. An input image is passed through a shared encoder to generate a set of features at 4 different scales relative to the input. We compute the initial predictions using each feature scale and then upsample all task-specific feature maps to the highest scale and combine them in the cross-scale fusion blocks. Finally, the output of each task-specific cross-scale fusion is passed as input to the CTAL module, where the features are refined and then processed by task-specific decoders to obtain the final predictions.} 
    \label{fig:emanet_ms}
\end{figure*}

\s{The prominent research directions for modern MTL methods in dense prediction explore either the optimization strategy~\cite{chen2018gradnorm,liu2022auto,xin2022current} or the design of the deep multitask architecture~\cite{misra2016cross,gao2019nddr,sinodinos2022attentive}. Within the vein of deep architecture design, Vandenhende \etal~\cite{9336293} further fragment the designs into encoder-focused and decoder-focused architectures. Their comprehensive MTL survey~\cite{9336293} highlights that for dense prediction, decoder-focused methods achieve greater performance than purely encoder-focused methods because cross-task patterns have more of an influence on the final pixel-wise predictions when propagated in the decoder. As the name suggests, decoder-focused models employ cross-task talk mechanisms within the decoder that allow for explicitly capturing and propagating inter-task patterns. An example of a pattern between tasks would be the alignment of segmentation edges with discontinuities in depth values. By providing a carefully designed approach to capture these inter-task relationships, decoder-focused architectures consistently achieve state-of-the-art performance in MTL~\cite{xu2018pad,zhang2019pattern,vandenhende2020mti}. Additionally, decoder-focused methods are significantly easier to develop and train because they can be fine-tuned using any off-the-shelf pretrained encoder; whereas encoder-focused methods need cross-task mechanisms to be specifically designed for a given backbone architecture and likely have to be trained from scratch. Given their performance and ease of use, decoder-focused methods have been regarded as the prevailing research direction for multitask architecture design~\cite{9336293}.}

The main research directions for modern MTL in dense prediction focus on either optimization strategies~\cite{chen2018gradnorm,liu2022auto,xin2022current} or deep multitask architecture design~\cite{misra2016cross,gao2019nddr,sinodinos2022attentive}. Vandenhende \etal~\cite{9336293} categorize architectures into encoder- and decoder-focused designs, highlighting that decoder-focused methods outperform encoder-focused ones in dense prediction because cross-task patterns have a stronger influence on pixel-wise predictions when propagated in the decoder. Decoder-focused models incorporate mechanisms within the decoder to explicitly capture and propagate inter-task patterns, such as the alignment of segmentation edges with depth discontinuities. This targeted approach enables these architectures to consistently achieve state-of-the-art MTL performance~\cite{xu2018pad,zhang2019pattern,vandenhende2020mti}. Furthermore, decoder-focused methods are easier to develop and train, as they can be fine-tuned using off-the-shelf pretrained encoders, unlike encoder-focused methods that require specially designed cross-task mechanisms for a given backbone and often need to be trained from scratch. Their superior performance and simplicity have made decoder-focused methods the leading research direction for multitask architecture design~\cite{9336293}.

\s{The current state-of-the-art decoder-focused algorithms leverage a refinement process called ``task-prediction distillation''~\cite{xu2018pad,zhang2019pattern,vandenhende2020mti}. The general idea behind this process involves improving initial task predictions by distilling cross-task pattern information from these initial predictions to generate improved final predictions. Specifically, this involves first using preliminary decoders to generate initial predictions for each task, then extracting the features from the penultimate layer of each of the task-specific decoders, augmenting the features with a cross-task distillation algorithm, and finally passing the augmented features through another set of decoders to generate the final task predictions.} 

State-of-the-art decoder-focused algorithms use a refinement process called ``task-prediction distillation"~\cite{xu2018pad,zhang2019pattern,vandenhende2020mti}. This process improves initial task predictions by distilling cross-task pattern information to produce better final predictions. It begins with preliminary decoders generating initial predictions for each task. Features are then extracted from the penultimate layer of these decoders, augmented using a cross-task distillation algorithm, and passed through another set of decoders to produce the final task predictions.

\s{PAD-Net~\cite{xu2018pad} is the first work to popularize this concept of task-prediction distillation for dense prediction in MTL. This framework is very similar to the one illustrated in ~\cref{fig:emanet_ms}, with the main difference being that PAD-Net used a single-scale feature extractor. Therefore, they only made initial predictions at a single scale (i.e., 1/4 input scale), and did not require a ``cross-scale distillation" phase, which involves combining features across multiple feature scales. PAD-Net also introduced the first ``cross-task distillation" module which is based on a convolutional self-attention algorithm. Although their cross-task distillation module can explicitly capture local patterns intra- and inter-task, it fails to explicitly capture long-range dependencies between the features of the initial task predictions. The details of this are discussed in ~\cref{method}.}

PAD-Net~\cite{xu2018pad} was the first to popularize task-prediction distillation for dense prediction in MTL. Its framework closely resembles that in ~\cref{fig:emanet_ms}, with the key difference being that PAD-Net employed a single-scale feature extractor. Consequently, it only made initial predictions at a single scale (1/4 input scale) and did not require a ``cross-scale distillation" phase to combine features across multiple scales. PAD-Net also introduced the first ``cross-task distillation" module based on a convolutional self-attention algorithm. While this module explicitly captures local intra- and inter-task patterns, it falls short in capturing long-range dependencies between features of the initial task predictions. Further details are discussed in ~\cref{method}.

\s{PAP-Net~\cite{zhang2019pattern} is another task-prediction distillation method that aims to overcome the shortcomings of PAD-Net's cross-task distillation module by explicitly modelling all local and long-range dependencies. To accomplish this, they  used a different self-attention algorithm in the cross-task distillation module which involves constructing a similarity matrix that contains similarity scores for every pair of features from a given initial task prediction. They referred to this similarity matrix as a ``task affinity matrix". The details of the self attention algorithm used to construct these matrices are discussed in ~\cref{method}. Although these task affinity matrices capture all local and long-range dependencies intra-task, the main problem with PAP-Net is its inter-task modelling. Specifically, they simply combine each task affinity matrix using a weighted sum, i.e., a learnable single weight per affinity matrix, and then diffuse this similarity information into the features of each initial task prediction via matrix multiplication. A problem with this approach is that learning a single weight (scalar) per affinity matrix suggests that all pairwise similarity patterns in the feature space are equally important. This simple cross-task mechanism was likely used because processing these matrices can be expensive, especially at larger feature scales. However, breakthroughs in attention mechanisms \cite{vaswani2017attention} have shown that cross-feature relationships are varied throughout the feature space, which makes it reasonable to assume this should also be the case for cross-task feature relationships. Consequently, there is substantial untapped potential in task affinity representations that can be leveraged to address the \textbf{absence of a cross-task distillation module that can model local and long-range dependencies intra- and inter-task. (i)}}

PAP-Net~\cite{zhang2019pattern} addresses the limitations of PAD-Net's cross-task distillation by explicitly modeling both local and long-range dependencies. It achieves this by using a different self-attention algorithm in the cross-task distillation module, constructing a similarity matrix—referred to as a ``task affinity matrix"—which contains similarity scores for each pair of features from the initial task prediction. The details of this algorithm are discussed in ~\cref{method}. While the task affinity matrices effectively capture intra-task local and long-range dependencies, PAP-Net's weakness lies in inter-task modeling. Specifically, it combines each task affinity matrix using a simple weighted sum, with a single learnable weight per matrix, and then diffuses this similarity information into the features via matrix multiplication. This approach assumes that all pairwise similarity patterns are equally important, which is problematic. Although this method likely simplifies the computation of these matrices, especially at larger scales, advances in attention mechanisms~\cite{vaswani2017attention} suggest that cross-feature relationships vary throughout the feature space, making it reasonable to expect this variation in cross-task relationships as well. Consequently, there is substantial untapped potential in task affinity representations that can be leveraged to address the \textbf{absence of a cross-task distillation module that can model local and long-range dependencies intra- and inter-task. (i)}
% , which CTAL addresses by properly capturing the nuances of cross-task similarity patterns throughout the feature space.

\s{A more recent work, MTI-Net~\cite{vandenhende2020mti}, shows that tasks with high affinities at a certain feature scale are not guaranteed to have high affinities at different scales. Consequently, they model task interactions at multiple scales to perform ``multi-scale task-prediction distillation". This is very similar to the multi-scale framework presented in ~\cref{fig:emanet_ms}, except in MTI-Net, they perform the ``cross-scale distillation" step after the ``cross-task distillation" step. Consequently, their framework requires preliminary decoders for every task at every feature scale, and a cross-task distillation module for every feature scale. Although this method yield performance improvements for multi-scale feature extractors, it comes at the cost of substantially more parameters. In our experiments using only 3 tasks with a multi-scale CNN feature extractor, MTI-Net had over double the model parameters of PAD-Net despite using the same cross-task distillation algorithm. These added model parameters make this approach susceptible to overfitting on simpler datasets and make the compute requirements scale poorly with more tasks. Therefore, we believe that \textbf{there is a need for a more parameter-efficient framework that can still leverage the benefits of multiscale processing (ii)}.}

More recently, MTI-Net~\cite{vandenhende2020mti} demonstrates that tasks with high affinities at one feature scale may not have high affinities at others. To address this, they model task interactions across multiple scales using ``multi-scale task-prediction distillation." This approach is similar to the multi-scale framework in ~\cref{fig:emanet_ms}, but in MTI-Net, the ``cross-scale distillation" step occurs after the ``cross-task distillation" step. As a result, their framework requires preliminary decoders and cross-task distillation modules for each task at every scale. While this improves performance for multi-scale feature extractors, it significantly increases the number of parameters. In our experiments with 3 tasks using a multi-scale CNN, MTI-Net had more than double the parameters of PAD-Net, despite using the same cross-task distillation algorithm. This parameter increase makes the approach prone to overfitting on simpler datasets and leads to poor scalability as the number of tasks grows. Therefore, we believe \textbf{there is a need for a more parameter-efficient framework that can still leverage the benefits of multi-scale processing (ii)}.

\s{Despite MTI-Net~\cite{vandenhende2020mti} being the most recent task prediction distillation method, it has only been evaluated using CNN backbones. Since the emergence of the Vision Transformer (ViT)~\cite{dosovitskiy2020image}, these high-performing decoder-focused algorithms haven't been investigated thoroughly with transformer backbones in the MTL literature. Instead, there's only been the emergence of encoder-focused multitask transformer methods, like InvPT~\cite{invpt2022} and TaskPrompter~\cite{taskprompter2023}. Similar to us, they argue that current multitask attention mechanisms in task prediction distillation models have limited scope in their cross-task pattern modelling. Their multitask attention methods can capture local and long-range relationships intra- and inter-task, but mainly because they operate on more compressed features as a consequence of using feature extractors 
with significantly higher parameter budgets. }

Despite MTI-Net~\cite{vandenhende2020mti} being the latest task-prediction distillation method, it has only been tested with CNN backbones. With the introduction of the Vision Transformer (ViT)~\cite{dosovitskiy2020image}, these high-performing decoder-focused algorithms have yet to be thoroughly explored using transformer backbones in the MTL literature. Instead, encoder-focused multitask transformer methods like InvPT~\cite{invpt2022} and TaskPrompter~\cite{taskprompter2023} have emerged. Like us, they argue that current multitask attention mechanisms in task-prediction distillation models have a limited ability to model cross-task patterns. Their attention mechanisms capture both local and long-range intra- and inter-task relationships, largely due to operating on more compressed features, a result of using feature extractors with substantially higher parameter budgets.

% However, \textbf{modelling local, global, and cross-task relationships has yet to be accomplished using \s{lightweight what do you mean by lightweight you mean CNN so give example} task prediction distillation techniques (iii).}
% When equipped with comparable Transformer backbones, the current CNN-based attention methods perform very similarly to the InvPT and TaskPrompter~\cite{taskprompter2023}. Consequently, we want to move away from the trend of increasing model capacity, and believe that \textbf{there is a need for a cross-task attention mechanism that can model local and global task interactions intra- and inter-task for CNN-based architectures (iii)}.}

\s{Explain WHY decoder-focused methods are worth more exploration as the recent methods just mentioned above are encoder-focused. }
\s{Another major issue with task-prediction distillation methods in light weight regimes (i.e., CNN-based models), has been their resource requirements. The increase in parameters is largely due to the size of the intermediate decoders and the distillation modules relative to the CNN backbone. The increase in floating point operations (FLOPs) is attributed to the additional parameters but mostly to the fact that CNN backbones tend to output features at high scales relative to the input, which can make the cross-scale and cross-task distillation processes very cumbersome. Transformer backbones on the other hand are typically much larger, so the relative increase in parameters from these task-prediction distillation techniques would be significantly less pronounced. Also, transformer backbones output features at much smaller scales relative to the input, which would drastically reduce the the number of FLOPs introduced to distill these features. Therefore, it seems that using a transformer backbone in a task-prediction distillation framework would offer a boost in performance with negligible additional resource consumption compared to traditional multitask learning. \textbf{However, there has yet to be any evaluation of these techniques using transformer-based backbones (iii)}.}

Another key challenge for task-prediction distillation methods in lightweight regimes (i.e., CNN-based models) has been their resource demands. The increase in parameters primarily stems from the size of intermediate decoders and distillation modules relative to the CNN backbone. The rise in floating point operations (FLOPs) is due not only to additional parameters but also to CNN backbones' tendency to output high-scale features, which makes the cross-scale and cross-task distillation processes very cumbersome. Transformer backbones, though typically larger, would experience a much smaller relative increase in parameters from task-prediction distillation techniques. Moreover, transformer backbones output features at smaller scales, significantly reducing the FLOPs required for distillation. As a result, applying a transformer backbone in a task-prediction distillation framework could boost performance with minimal additional resource consumption compared to traditional multitask learning. \textbf{However, these techniques have yet to be evaluated with transformer-based backbones (iii)}.

\s{Despite their success, current decoder-focused methods have yet to address (i), (ii), (iii). We address all of these issues by introducing our novel Cross-Task Affinity Learning (CTAL) module for improved task prediction distillation. CTAL strategically aligns the task affinity matrices using careful reshaping and interleaved concatenations, which allows us to leverage grouped convolutions to realize massive reductions in model parameters compared to standard convolutions. Additionally, since we exhaustively model every pairwise feature relationship within a task and across all tasks, we can use these grouped convolutions without fear of information loss. This effectively solves issue (i), as we can tap into the full potential of affinity matrix representations to explicitly model all pairwise interactions intra- and inter-task using very few additional model parameters. In fact, we can outperform STL baselines with both CNN and transformer backbones using less than half the number of parameters. Our method also extends to a multiscale framework by applying deep supervision to initial task predictions at multiple scales. However, unlike MTI-Net, we fuse the initial task predictions from every scale and then perform cross-task distillation using a single module, as seen in ~\cref{fig:emanet_ms}. Consequently, our multi-scale framework is parameter efficient and outperforms MTI-Net using 12.9\% fewer model parameters; which addresses issue (ii). Therefore, we can summarize our contributions as follows:}

Despite their success, current decoder-focused methods have not fully addressed (i), (ii), and (iii). We tackle these issues with our novel Cross-Task Affinity Learning (CTAL) module for enhanced task prediction distillation. CTAL aligns task affinity matrices through careful reshaping and interleaved concatenations, enabling the use of grouped convolutions to achieve significant reductions in model parameters compared to standard convolutions. By exhaustively modeling every pairwise feature relationship within and across tasks, we leverage grouped convolutions without risking information loss, effectively solving issue (i). This approach fully utilizes affinity matrix representations, allowing us to explicitly model all pairwise interactions intra- and inter-task with minimal additional parameters. Notably, we outperform STL baselines with both CNN and transformer backbones using less than half the parameters. Our method also extends to a multi-scale framework, applying deep supervision to initial task predictions across multiple scales. Unlike MTI-Net, we fuse the initial predictions from all scales before performing cross-task distillation using a single module, as shown in ~\cref{fig:emanet_ms}. This parameter-efficient multi-scale framework outperforms MTI-Net with 12.9\% fewer parameters, addressing issue (ii).

In summary, our contributions are as follows:
\s{I would like to see a stronger emphasis on the novel aspects of our work. Currently, they seem somewhat underrepresented. For example, you could highlight that this is the \textbf{first} MTL method capable of fully processing affinity matrices, which sets it apart from existing approaches. Rather than writing 'a method,' assert that it is the 'first' where applicable. This is the section where you need to convince reviewers of the work's novelty, but that message didn't come through clearly when I read it.}
\\
\textbf{1.} A novel cross-task distillation module (CTAL) that is the first method to fully process task affinity matrices in a parameter-efficient manner to exhaustively model all local and long-range dependencies intra- and inter-task. This addresses (i);
\\
\textbf{2.} A novel light-weight multi-scale framework that yields the benefits of multi-scale deep supervision, while only needing a single scale for cross-task distillation. This addresses (ii).
\\
\textbf{3.} The first thorough evaluation of task prediction distillation methods for both CNN and ViT-based backbones; which also demonstrates that task-prediction distillation is very well suited for MTL solutions using modern backbones. This addresses (iii).
\s{I would also suggest adding an item mentioning that the method is applied to three popular benchmark MTL datasets, where it achieves state-of-the-art performance in terms of [specific metrics]. Highlight that our method accomplishes this with only a fraction of the learnable parameters compared to single-task learning. Including specific numbers would make this even more eye-catching. The same adjustment should be made in the Abstract section to strengthen the impact.}
%
% \begin{enumerate}
%     \item A method to process affinity matrices in a parameter-efficient manner while exhaustively modelling all intra- and inter-task relationships in CNN-based architectures. This addresses (i, iii).
%     \item A light-weight multiscale framework that yields the benefits of multiscale deep supervision, while only needing a single scale for task-prediction distillation. This addresses (ii).
% \end{enumerate}

As a result of our contributions, we achieve significant multitask performance improvements using a fraction of the learnable parameters compared to single-task learning.

\s{I like the below discussion shown in Olive. I would like to see this in the Introduction, as it directly hits the target! Besides, this is a discussion about other methods and should not be under Our method.}
\begin{figure*}[ht!]
    \centering
    \includegraphics[width=0.95\linewidth]{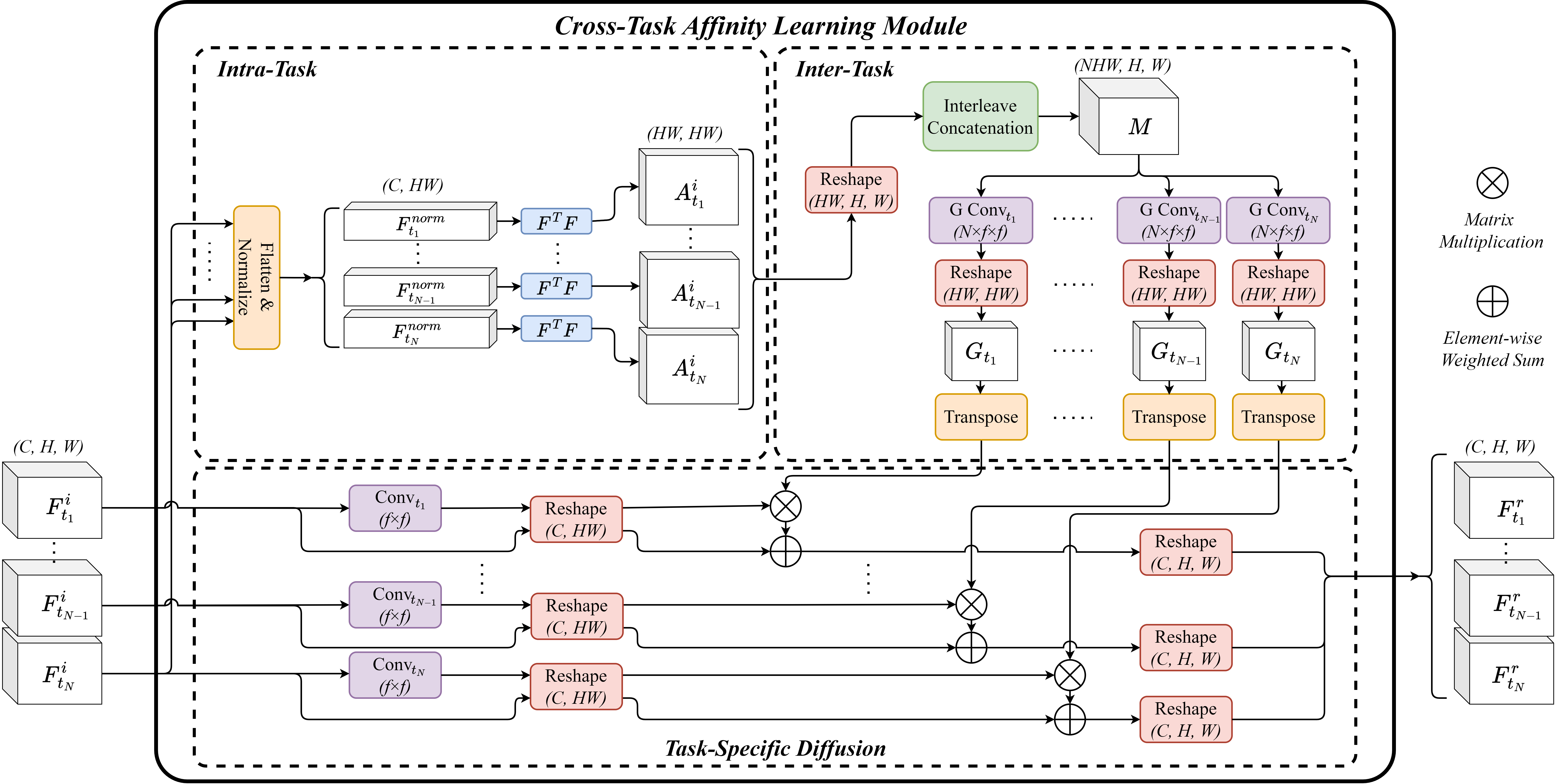}
    \caption{A diagram of the Cross-Task Affinity Learning (CTAL) module that is comprised of three stages: Intra-Task, Inter-Task, and Task-Specific Diffusion. We compute the Gram matrix of the flattened and normalized views of the initial task prediction features $\bm{F}^i_{t_k}$ to obtain the task-specific affinity matrices $\bm{A}^i_{t_k}$. We then reshape $\bm{A}^i_{t_k}$ to the original spatial dimensions and perform an interleaved concatenation of all $HW$ channels for each task to obtain the joint affinity matrix $\bm{M}$. Each of the $HW$ sets of $N$ channels is processed by a task-specific grouped convolution ($\text{G Conv}_{t_k}$) and then diffuses its information to a projected view of $\bm{F}^i_{t_k}$ via matrix multiplication and an element-wise weighted sum to obtain the final refined features $\bm{F}^r_{t_k}$.}
    \label{fig:ctal}
\end{figure*}
\s{If you moved above olive to Intro, then here provide a short opening paragraph mentioning that we will explain our method in the below 4 sections... }

\s{I SUGGEST To have a header as Proposed Method with two subsections: CTAL and Multi-Scale CTAL? }
\section{Proposed Method}
\label{method}
\s{As mentioned prior, the main difference between PAD-Net and PAP-Net is the type of self-attention algorithm they use in their cross-task distillation modules. Generally speaking, in the dense scene prediction literature, attention maps are generated slightly differently than the traditional methods for low dimensional token embeddings~\cite{vaswani2017attention}. Specifically, there are two prevailing approaches to applying an attention map to a set of image features. The first approach involves using convolutional blocks to process the features and obtain an attention mask with the same shape. The attention mask undergoes an activation (i.e., Sigmoid) to set all values to be between 0 and 1. The resulting attention map is applied to the features via element-wise multiplication, which has been applied in STL~\cite{woo2018cbam}, and MTL~\cite{liu2019end,xu2018pad}. We will refer to this as \b{``element-wise multiplication attention''} (EM attention), which is used by PAD-Net. The second approach aims to explicitly model long-range dependencies of features by computing the Gram matrix (inner products of all pairs of column vectors) of the features after flattening them along the spatial dimensions. The corresponding matrix (i.e., affinity matrix) is then diffused to the original features via matrix multiplication. We will refer to this method as \b{``matrix multiplication attention''} (MM attention), which is used by PAP-Net. MM attention has also been used in STL~\cite{fu2019dual}, and MTL~\cite{zhang2019pattern}. In MTL, the EM attention from PAD-Net and the MM attention from PAP-Net achieve almost identical results. However, PAP-Net uses considerably fewer model parameters at the cost of more \s{floating point operations (}\b{FLOPs} at higher feature scales. Their comparable performance is likely attributed to the fact that the affinity matrices explicitly capture long-range dependencies between every feature pair and diffuse them across all features using MM attention; whereas EM attention approaches only learn long-range dependencies implicitly by training convolutional filters that process local patches throughout the entire spatial dimension.}

As mentioned earlier, the primary difference between PAD-Net and PAP-Net lies in the self-attention algorithm used in their cross-task distillation modules. In dense scene prediction, attention maps are generated differently from traditional methods for low-dimensional token embeddings~\cite{vaswani2017attention}. There are two main approaches for applying attention maps to image features. The first uses convolutional blocks to process features and produce an attention mask of the same shape. The mask undergoes activation (e.g., Sigmoid) to constrain values between 0 and 1, and the attention map is applied to the original features via element-wise multiplication. This method, referred to as element-wise multiplication attention (EM attention), is used by PAD-Net and has been applied in both STL~\cite{woo2018cbam} and MTL~\cite{liu2019end,xu2018pad}. The second approach explicitly models long-range dependencies by computing the Gram matrix (i.e., inner products of all column vector pairs) of the features after flattening them spatially. This matrix, called an affinity matrix, is diffused to the original features via matrix multiplication, referred to as matrix multiplication attention (MM attention), as used by PAP-Net and applied in STL~\cite{fu2019dual} and MTL~\cite{zhang2019pattern}. Both EM attention (PAD-Net) and MM attention (PAP-Net) perform similarly in MTL, though PAP-Net requires fewer parameters at the cost of more FLOPs at higher feature scales. This comparable performance likely stems from MM attention's explicit modeling of long-range dependencies between all feature pairs, while EM attention implicitly learns these dependencies by training convolutional filters on local patches across the spatial dimension.

\s{The fact that MM attention in PAP-Net achieves almost identical performance to EM attention using a simple weighted sum for cross-task fusion is astonishing. We believe that there is substantial untapped potential in these affinity representations, and therefore, we were inspired to develop a parameter-efficient cross-task attention mechanism to optimally model local and long-range interactions intra- and inter-task. For parameter efficiency, we aim to employ MM attention, which involves processing large affinity matrices. Therefore the challenge lies in minimizing the parameters introduced to process these matrices. We propose our CTAL distillation module to efficiently and exhaustively model all cross-task patterns; which is comprised of three stages: intra-task modelling, inter-task modelling, and task-specific diffusion. The details for each of these steps are provided in~\cref{ctal}.}

The fact that MM attention in PAP-Net achieves nearly identical performance to EM attention, despite using a simple weighted sum for cross-task fusion, is remarkable. We believe there is significant untapped potential in affinity representations, which led us to design a parameter-efficient cross-task attention mechanism to effectively model local and long-range interactions both intra- and inter-task. To ensure parameter efficiency, we focus on MM attention, which processes large affinity matrices. The main challenge is minimizing the parameters required for this processing. Our proposed CTAL distillation module efficiently models all cross-task patterns and consists of three stages: intra-task modeling, inter-task modeling, and task-specific diffusion. Details for each step are provided in~\cref{ctal}.

For the purposes of reducing the computational burden of the multi-scale task-prediction distillation framework proposed by MTI-Net, we propose an alternative framework that removes the need for performing cross-task distillation at multiple scales. The details of this framework are described in \cref{multiscale} 

\subsection{CTAL}
\label{ctal}

\subsubsection{Intra-Task Modelling}
\s{I suggest before digging into formulations, give a short motivation for why we are doing this and what are the expectations of this section. same for the other subsections. The current writing implies that we are doing some vague reshapings for a reason that is not clearly defined.  }
The purpose of this stage is to effectively model all local and long-range dependencies within the features of each initial task prediction individually. This is the same procedure used by PAP-Net to generate the task affinity matrices, as described in ~\cref{sec:intro}.

As seen in~\cref{fig:ctal}, for a given task $t_k$ \{$k \in [1,N]$\}, we follow the standard procedure to generate the affinity matrices, $\bm{A}^i_{t_k} \in \mathbb{R}^{HW,HW}$, where $N$ is the number of tasks. \s{Define N, H and W C } This involves taking the features of the initial predictions, $\bm{F}^i_{t_k} \in \mathbb{R}^{C,H,W}$, flattening the spatial dimensions, \s{i.e. W and H??, }performing L2 normalization for each column, and computing the Gram matrix (the inner products of all pairs of column vectors), where $C$, $H$, and $W$ are the channel, width, and height dimensions respectively. Throughout this paper, the superscript $i$ refers to initial predictions. Each row in $\bm{A}^i_{t_k}$ contains the cosine similarities of a feature, $x_{u,v} \in \bm{F}^i_{t_k}$, with every other feature in $\bm{F}^i_{t_k}$, where $x_{u,v}$ is a $C$ dimensional vector located at coordinates $u \in [1,H]$ and $v \in [1,W]$.

\subsubsection{Inter-Task Modelling}
\s{Same as above, first explain why. then how? }
The purpose of this stage is to take the individual task affinity matrices and combine them in a way that effectively captures all the relevant feature patterns across tasks. This stage is where PAP-Net implemented a simple weighted sum to combine the affinity matrices. Instead, we take a novel approach by learning a detailed task-specific attention map across all affinity matrices.

\begin{figure}[t!]
    \centering    \includegraphics[width=\linewidth]{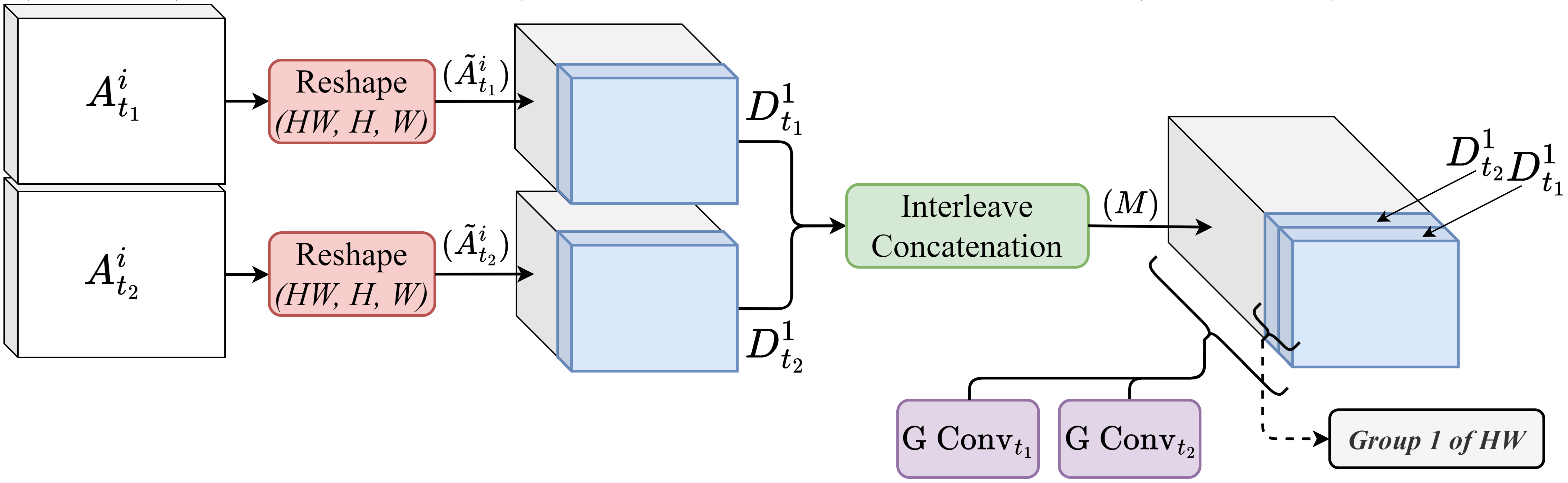}
    \caption{An illustration of the interleave concatenation procedure used to align the channels for grouped convolutions in a two-task scenario. \s{M is the whole block, right?. Currently, it seems it refers to only one slot of the block. you may revise the curly bracket of M to expand over the depth of the block? }}
    \label{fig:conv-groups}
\end{figure}

First, we reshape $\bm{A}^i_{t_k}$ into $\bm{\tilde{A}}^i_{t_k} \in$ $\mathbb{R}^{HW,H,W}$. This restores the original spatial dimensions of the features, but now the $HW$ channels at a given 2-dimensional position $(u, v)$ contain the cosine similarities of the feature $\s{x_{u,v}}$ with all other features. Therefore, in this configuration, for example, the entire first channel of $\bm{\tilde{A}}^i_{t_k}$, i.e., $D^1_{t_k}$ like in~\cref{fig:conv-groups}, corresponds to the similarities of all $x_{u,v}$ with $x_{1,1}$, which is also aligned at position $(u,v)$. This is a useful property for maintaining spatial coherence during subsequent processing.

Next, to fuse the reshaped affinity matrices, $\bm{\tilde{A}}^i_{t_k}$, we first perform an interleave concatenation operation. As seen in~\cref{fig:conv-groups}, this involves concatenating the first channel of each $\bm{\tilde{A}}^i_{t_k}$ (i.e. $D^1_{t_1}$ and $D^1_{t_2}$), and then the second channels, and so on for all $HW$ channels of each $\bm{\tilde{A}}^i_{t_k}$. This gives us the joint affinity matrix, $\bm{M} \in \mathbb{R}^{NHW,H,W}$\s{, where $N$ is the number of tasks in the MTL system}. Depending on the spatial dimensions of the data, this $\bm{M}$ can be very large, so processing it using standard convolutions would be very expensive. Instead, we aim to leverage parameter-efficient grouped convolutions~\cite{xie2017aggregated,chollet2017xception,sandler2018mobilenetv2}. A grouped convolution splits input channels into groups, applies separate convolutions to each group, and then concatenates the outputs; which significantly reduces parameters and computational cost. The grouped convolutional blocks in~\cref{fig:ctal} and~\cref{fig:conv-groups} are labeled ``G Conv".

\s{Here, quickly explain what G-convolution is and its benefits. }

The way we strategically organized $\bm{M}$ strongly justifies the use of grouped convolutions to perform our multitask fusion without the fear of losing important cross-task information. This is because every group of $N$ channels already contains task interactions between a given feature $x_{u,v}$ with every other feature across all tasks. So not only does this significantly reduce the number of model parameters required to fuse $\bm{M}$ for every task, but it also allows us to learn $HW$ spatially coherent kernels that specifically focus on learning the relationships of a given feature $x_{u,v}$ with all other features across all tasks. \s{Repeat again why this is important to learn the relationships of given features... throughout the method section, keep emphasizing why it is important and how we realize that-- this is what matters in this paper because the remaining looks like only playing with the dimensions ... ... }When considering a traditional convolution on $\bm{M}$, we would require $HW$ kernels of size $NHW\times f\times f$, whereas our approach only requires $HW$ kernels of size $N\times f \times f$, where $f$ is the size of the convolutional filter. This translates to $HW$ times fewer parameters used. For example, working with 3 tasks at $72\times96$ feature size and $f=3$, for the NYUv2 dataset~\cite{Silberman:ECCV12}, we use only 187K parameters compared to the 1.29B parameters needed for a single standard convolutional layer \s{required in methods like ....}.

After processing the shared $\bm{M}$ for every task, we obtain $N$ matrices and reshape each of them back into $HW\times HW$ to obtain each $\bm{G}_{t_k} \in \mathbb{R}^{HW,HW}$. Now, in a given $\bm{G}_{t_k}$, each row contains information about the relationship of a single feature with every other feature across all tasks. Since we use $f\times f$ kernels where $f>1$, we also embed cross-task spatial interaction patterns. Next, we transpose the matrix so that the row containing all pertinent information for $x_{u,v}$ is stored at location $(u,v)$ after the diffusion process, which perfectly maintains spatial coherence throughout the entire attention process. \s{repeat why it is important to have such property}

\subsubsection{Task-Specific Diffusion}
\s{again, first why, then how. why do we need diffused features ...}
The purpose of this final stage is to take the learned cross-task affinity attention mappings, $\bm{G}_{t_k}$, and diffuse them back into the features of the initial task predictions, $\bm{F}^i_{t_k}$, to get the refined features, $\bm{F}^r_{t_k}$, that will be used to generate the final predictions.

The subsequent diffusion process uses MM attention, which as the name suggests, involves performing a matrix multiplication to obtain the diffused features, $\bm{F}^d_{t_k} \in \mathbb{R}^{C,HW}$:
\begin{equation}
    \bm{F}^d_{t_k} = \bm{F}^p_{t_k} \times \bm{G}_{t_k}^{\top},
\label{diffuse}
\end{equation}
%
% \s{a discussion and interpretation on what is the effect of such multiplication on every individual feature present in Fptk... it is not clear what happens if there is a similarity between two specific features within and between tasks... we had such a discussion before on the whiteboard.... This section should clearly explain in wording how we finally update features to reflect the effect of inter-intra task similarities... }
where $\bm{F}^p_{t_k} \in \mathbb{R}^{C,HW}$ is the reshaped convolution projection of $\bm{F}^i_{t_k}$. Through this matrix multiplication, every value in $\bm{F}^d_{t_k}$ is the result of the dot product between a row vector in $\bm{F}^p_{t_k}$ containing values from all $HW$ features and a column vector in $\bm{G}_{t_k}^{\top}$ containing cross-task affinity pattern information between a feature $x_{u,v}$ and all other $HW$ features, where $\top$ corresponds to the transpose operation. When there is a high-affinity pattern across tasks, the value of $x_{u,v}$ will become larger relative to other features with lower-affinity patterns, similar to the scaling behaviour from EM attention. As seen in Equation~\ref{blend}, the diffused features are then blended with the original features using element-wise addition with scalar weighing parameter $\gamma$ to obtain the refined features $\bm{F}^r_{t_k} \in \mathbb{R}^{C,H,W}$. This blending ensures the refined features do not deviate too far from the original features.
\begin{equation}
    \bm{F}^r_{t_k} = \gamma*\bm{F}^d_{t_k} + (1-\gamma)*\bm{F}^i_{t_k}
\label{blend}
\end{equation}

% \subsection{Channel CTAL}
% Applying CTAL to the channel dimensions is almost identical to how we applied it in the spatial context, but with a few modifications. First, we  compute the inner product along the HW dimension to obtain $A_{t_k} \in \mathbb{R}^{C,C}$. Since there is no notion of spatial coherence, we can keep the affinity matrices in their current shape. Next, to obtain the joint matrix $M \in \mathbb{R}^{NC,C}$, we perform an interleave concatenation along the rows, as seen in \ref{fig:conv-groups}. Due to this strategic alignment, we can once again leverage parameter efficient grouped convolutions (1D) using $N\times 1\times 3$ filters, where each filter specifically focuses on learning the relationships of a given channel, $c_{d}$ where $d \in [1,C]$, with all other channels across all tasks. The subsequent diffusing process is the same where we perform a matrix multiplication with the appropriately shaped initial features to obtain the diffused features; which are then blended with the initial features using \ref{blend}.

% Using this grouped convolutional approach in the channel context, we only need to use $C$ kernels of size $N\times 1\times 3$, as opposed to using kernels of size $(NC)\times 1\times 3$ for regular convolutions. This leads to using $C$ fewer model parameters. In our 3 task setting for NYUv2 using $C=128$, we use only 

\subsection{Multi-Scale Framework}
\label{multiscale}
\s{Again, make sure of first why? then how.}

\s{~\cref{fig:emanet_ms} is an overview of the general task-prediction distillation framework we propose. Using a single scale feature extractor, there would only be an input, a backbone (feature extractor), the initial predictions, a cross-task distillation module, and then the final predictions. However, more recent works have shown the benefits of multi-scale processing~\cite{vandenhende2020mti}; which uses a multi-scale backbone and introduces a cross-scale distillation step. Consequently, we will experiment with two variants of the task-prediction distillation framework. The first variant performs cross-scale fusion after the feature extractor and makes initial predictions using a single feature scale (SS), while the other makes initial predictions at multiple feature scales (MS) and performs cross-scale fusion afterwards.\s{You may call them CTALss and CTALms -- you may even refer to them in the introduction section when introducing the ms version. } The latter is illustrated in~\cref{fig:emanet_ms}, which is used to compare with multi-scale distillation methods like MTI-Net~\cite{vandenhende2020mti}. However, our model is more efficient and scalable as we only need a single cross-task distillation module (i.e. CTAL) rather than having one module for every scale. We accomplish this by combining the initial prediction features from each scale with the cross-scale fusion (CSF) blocks prior to performing task prediction distillation in CTAL. For cross-scale fusion using CNN backbones, we follow the same procedure as~\cite{vandenhende2020mti}, which involves up-sampling all features to the 1/4 input scale, concatenating them along the channel dimension, and combining them through a convolutional block. For comparison with other methods that operate on a single feature scale, we perform the same cross-scale fusion on the multi-scale features generated by the shared encoder and generate only a single set of initial predictions.}

~\cref{fig:emanet_ms} provides an overview of our proposed task-prediction distillation framework. Using a single-scale feature extractor, the process would include an input, a backbone (feature extractor), initial predictions, a cross-task distillation module, and the final predictions. However, recent works have highlighted the advantages of multi-scale processing~\cite{vandenhende2020mti}, which uses a multi-scale backbone and introduces a cross-scale distillation step. Thus, we experiment with two variants of the framework: the first performs cross-scale fusion after the feature extractor, generating initial predictions using a single feature scale (SS), while the second makes initial predictions at multiple scales (MS) and performs cross-scale fusion afterward. The latter, illustrated in~\cref{fig:emanet_ms}, is used for comparison with multi-scale distillation methods like MTI-Net~\cite{vandenhende2020mti}. Our model, however, is more efficient and scalable, requiring only a single cross-task distillation module (CTAL) instead of one for each scale. We achieve this by combining initial prediction features from each scale with cross-scale fusion (CSF) blocks before performing task prediction distillation in CTAL. For cross-scale fusion with CNN backbones, we follow~\cite{vandenhende2020mti}'s procedure, up-sampling all features to the 1/4 input scale, concatenating them along the channel dimension, and combining them with a convolutional block. To compare with single-scale methods, we apply the same cross-scale fusion to multi-scale features generated by the shared encoder and produce only a single set of initial predictions.

\section{Experimental Setup}
\label{setup}
\subsection{Datasets}
\label{datasets}
\s{We perform our experiments on NYUv2~\cite{Silberman:ECCV12}, Cityscapes~\cite{Cordts2016Cityscapes}, and PASCAL-Context~\cite{mottaghi_cvpr14} datasets, which are all very popular for multitask learning. \textbf{NYUv2} contains 1449 densely labelled RGB-depth images of indoor scenes. The tasks associated with this dataset include semantic segmentation, depth estimation, and surface normals. \textbf{Cityscapes} is a larger dataset containing 3475 outdoor urban street scenes with fine annotations taken from 50 cities over several months of the year. The tasks associated with this dataset are semantic segmentation and depth estimation. \textbf{PASCAL-Context}~\cite{mottaghi_cvpr14} is an even larger dataset derived from the PASCAL VOC 2010 challenge \cite{pascal-voc-2010}, containing pixel-wise annotations for 10,103 images. These images cover a wide range of indoor and outdoor scenes with various objects. The tasks associated with this dataset are semantic segmentation, human parts segmentation, saliency detection, edge detection, and surface normals.

For both NYUv2 and Cityscapes, we use publicly available preprocessed datasets courtesy of~\cite{liu2019end}. For PASCAL-Context, we use the data and preprocessing code from~\cite{bruggemann2021exploring}. More details on these datasets can be found in the appendix.}

We conduct our experiments on the widely-used NYUv2~\cite{Silberman:ECCV12}, Cityscapes~\cite{Cordts2016Cityscapes}, and PASCAL-Context~\cite{mottaghi_cvpr14} datasets, popular in multitask learning. \textbf{NYUv2} contains 1449 densely labeled RGB-depth indoor scene images, with tasks including semantic segmentation, depth estimation, and surface normals. \textbf{Cityscapes} is a larger dataset with 3475 outdoor urban street scenes, annotated from 50 cities, supporting semantic segmentation and depth estimation tasks. \textbf{PASCAL-Context}~\cite{mottaghi_cvpr14}, derived from the PASCAL VOC 2010 challenge~\cite{pascal-voc-2010}, contains pixel-wise annotations for 10,103 images of diverse indoor and outdoor scenes. Its tasks include semantic segmentation, human parts segmentation, saliency detection, edge detection, and surface normals. We use publicly available preprocessed datasets from~\cite{liu2019end} for NYUv2 and Cityscapes, and from~\cite{bruggemann2021exploring} for PASCAL-Context. More details on these datasets are provided in the appendix.

\subsection{Tasks and Performance Metrics}
\s{Always assume that the reviewers are not experts in MTL. So tell them below is std to the field by providing Refs. }
\s{CAN you give a ref for each of the below parts? Just to show these are common metrics to the field}
\s{THIS section is too long. I wonder if it can be shortened? }

Following the MTL literature~\cite{taskprompter2023,invpt2022,vandenhende2020mti,liu2019end}, Semantic segmentation (SemSeg) and human parts segmentation (HPSeg) are evaluated using mean intersection over union (mIoU). Depth estimation (Depth) is evaluated using relative depth error (relErr). Surface normals prediction (Normals) is evaluated using the mean error (mErr). Saliency detection (Sal) is evaluated using the max F-measure (maxF). Edge detection (Edge) is evaluated using the binary cross-entropy loss (Loss) on the validation set. Finally, MTL Gain~\cite{maninis2019attentive} is an aggregate measure of the overall multitask improvement of method $m$ with respect to a single task learning baseline $b$ for all tasks $t \in [1,N]$, as seen in Equation \ref{deltam}.
\begin{equation}
    \Delta_m = \frac{1}{N}\sum_t^N(-1)^{l_t}(M_{m,t}-M_{b,t})/M_{b,t}
\label{deltam}
\end{equation}
where $l_t = 1$ if a lower value of metric $M$ is favorable, and 0 otherwise. In our results, the metrics where larger values are favourable are denoted with ($\uparrow$) and smaller values with ($\downarrow$). We will treat $\Delta_m$ as a percentage in our evaluation. More details about each task and their metrics can be found the appendix.

\subsection{Baselines}
\s{It is standard practice in MTL to compare against the traditional STL and MTL baselines. The STL baseline involves using a single network for each task, where each network uses a comparable backbone and output heads as the proposed model for a fair comparison. The MTL baseline uses a hard parameter sharing network~\cite{9336293}, which involves sharing the backbone layers across all tasks, and then feeding the shared feature representation to each task-specific output head. Following~\cite{vandenhende2020mti} for the CNN models, we equip them with a high-resolution network backbone (HRNet18)~\cite{wang2020deep} to generate multi-scale features that are processed by scale-specific output heads and aggregated using the aformentioned cross-scale fusion blocks. For the transformer models, we will use a SwinV2 \cite{liu2022swin} backbone, which has shown state-of-the-art performance for dense vision tasks and already has multiscale feature extraction and aggregation built-in.

Since we are proposing a novel cross-task distillation module, we must compare with the current best approaches in this domain. Therefore, we will be evaluating against PAD-Net and PAP-Net, which will serve as the EM attention and MM attention baselines, respectively. Additionally, they also serve as our single-scale baselines. Next, we will also compare against MTI-Net, which will serve as our multi-scale baseline. Since the multi-scale framework requires a backbone that outputs features at multiple scales, it will not be considered during the evaluation with the transformer backbones. This is because current state-of-the-art ViT models for dense predictions, like SwinV2~\cite{liu2022swin}, use multi-scale feature extractors progressively throughout the encoding process and provide an aggregated feature representation at a single scale as a result. We deemed modifying the architecture of these ViT models to fit the multi-scale requirements be out-of-scope for this work.

All experiments for our models and the baselines are performed 3 times using a different seed for each experiment. The same set of 3 seeds is used across all models for consistency. The results in all tables contain the average of the converged values across all 3 experiments for each model. Results including the standard deviation across the 3 runs can be found in the appendix. All experiments using CTAL in a single-scale and multi-scale framework will be denoted by CTAL$_{SS}$ and CTAL$_{MS}$ respectively.}

In MTL, it is standard practice to compare against traditional STL and MTL baselines. The STL baseline uses a separate network for each task, with each network sharing the same backbone and output heads as the proposed model to ensure fair comparison. The MTL baseline uses a hard parameter sharing network~\cite{9336293}, where the backbone is shared across all tasks, and the shared feature representation is fed to task-specific output heads. Following~\cite{vandenhende2020mti}, CNN models are equipped with an HRNet18 backbone~\cite{wang2020deep} to generate multi-scale features, processed by scale-specific output heads and aggregated using cross-scale fusion blocks. For transformer models, we use a SwinV2-S backbone~\cite{liu2022swin}, which has demonstrated state-of-the-art performance in dense vision tasks and already incorporates multi-scale feature extraction and aggregation. As we are proposing a novel cross-task distillation module, we compare against the best current approaches in this domain. PAD-Net and PAP-Net serve as baselines for EM attention and MM attention, respectively, and also act as our single-scale baselines. Additionally, we compare against MTI-Net as our multi-scale baseline. Since the multi-scale framework requires a backbone that outputs features at multiple scales, it is excluded from the evaluation with transformer backbones. This is because state-of-the-art ViT models for dense predictions, like SwinV2~\cite{liu2022swin}, progressively extract multi-scale features throughout encoding and provide an aggregated single-scale feature representation. Modifying ViT models to meet multi-scale requirements is beyond the scope of this decoder-focused work. Additional implementation details for all models can be found in the appendix.

All experiments for our models and the baselines are performed three times, using a different seed for each run. The same set of three seeds is used across all models for consistency. The tables report the average of the converged values across all three experiments for each model. Results including the standard deviation across the three runs are available in the appendix. Models using CTAL in a single-scale and multi-scale framework are denoted as CTAL$_{SS}$ and CTAL$_{MS}$, respectively.

\section{Results}

% \s{have a subsection for hyperparameter sensitivity and show (maybe partially) the results here rather than in the appendix, if there is space available}

\subsection{Comparison to State-of-the-Art}

\s{This section must be before ablation. This is the most important result -- here, you impress reviewers by beating SOTA -- then later, you tell them the effect of each module in an ablation study.}

\s{You may briefly explain why MTI-net is not included in our experiments.}

% \begin{table}[htb!]
%     \centering
%     \resizebox{0.4\textwidth}{!}{%
%     \begin{tabular}{cccc}
%     \toprule
%       \multirow{2}{*}{Model} & \textbf{Sem. Seg.} & \textbf{Depth} & \textbf{MTL Gain}\\
%        & mIoU $\uparrow$ &  relErr $\downarrow$ & $\Delta_m \uparrow$\\
%      \midrule
%      STL & 48.89 & 29.91 & \hspace{5pt}+0.00\\ 
%      MTL & 49.78 & 31.80 & \hspace{5pt}-2.25 \\
%      \midrule
%      PAP-Net & 50.82 & 26.97 & \hspace{5pt}+6.89 \\
%      PAD-Net & 50.67 & 27.37 & \hspace{5pt}+6.07 \\
%      EMA-Net (SS) & \textbf{51.36} & \textbf{23.84} & \textbf{+12.67} \\
%      \midrule
%      MTI-Net & 51.77 & 29.90 & \hspace{5pt}+2.96 \\
%      EMA-Net (MS) & \textbf{51.94} & \textbf{22.89} & \textbf{+14.84} \\
%     \bottomrule
%     \end{tabular}
%     }
%     \caption{Validation set performance taken across all tasks on Cityscapes. Values in bold indicate the best value in a given column for multitask models in SS and MS configurations.}
%     \label{tab:sota_cityscapes}
% \end{table}

\begin{table}[htb!]
    \centering
    \setlength{\tabcolsep}{4pt}
    \resizebox{0.48\textwidth}{!}{%
    \begin{tabular}{c c c c c c c c c c} %{\textwidth}
    \toprule
      & &\multicolumn{4}{c}{\textit{NYUv2 (CNN)}} & & \multicolumn{3}{c}{\textit{Cityscapes (CNN)}}\\
      \cline{3-6}
      \cline{8-10}\\[-2ex]
      \multirow{2}{*}{Model} & &\textbf{SemSeg} & \textbf{Depth} & \textbf{Normals} & \multirow{2}{*}{$\Delta_m \uparrow$} & & \textbf{SemSeg} & \textbf{Depth} & \multirow{2}{*}{$\Delta_m \uparrow$}\\ 
      % & & & & & & & Angle Distance & \multicolumn{3}{c}{Within t\degree} \\ 
      % Model & \multicolumn{2}{c}{(Higher Better)} & & \multicolumn{2}{c}{(Lower Better)} & & (Lower Better) & \multicolumn{3}{c}{(Higher Better)} \\ 
      & & mIoU $\uparrow$ & relErr $\downarrow$ & mErr $\downarrow$ & & & mIoU $\uparrow$ & relErr $\downarrow$ & \\
     %\midrule
     \cline{1-1}\cline{3-6}\cline{8-10}\\[-2ex]
     STL & & 49.23 & 0.1636 & 23.15 & +0.00 & & 48.89 & 29.91 & \hspace{5pt}+0.00\\ 
     MTL & & 49.25 & 0.1658 & 24.16 & -1.89 & & 49.78 & 31.80 & \hspace{5pt}-2.25\\ 
      %\midrule
      \cline{1-1}\cline{3-6}\cline{8-10}\\[-2ex]
     PAD-Net & & 50.23 & 0.1622 & 23.63 & +0.27 & & 50.67 & 27.37 & \hspace{5pt}+6.07\\
     PAP-Net & & 50.00 & 0.1615 & 23.78 & +0.04 & & 50.82 & 26.97 & \hspace{5pt}+6.89\\
     %EMA-Net (C Att) & 49.40 & 73.07 & & 0.1609 & 0.3864 & & 23.48 & 33.34 & 69.71 & 71.72 & & +1.17\\ 
     CTAL$_{SS}$ & & \textbf{51.59} & \textbf{0.1607} & \textbf{22.84} & \textbf{+2.64} & & \textbf{51.36} & \textbf{23.84} & \textbf{+12.67}\\
     %\midrule
     \cline{1-1}\cline{3-6}\cline{8-10}\\[-2ex]
      MTI-Net & & 51.51 & 0.1538 & 23.50 & +3.04 & & 51.77 & 29.90 & \hspace{5pt}+2.96\\
     CTAL$_{MS}$ & & \textbf{52.70} & \textbf{0.1529} & \textbf{22.99} & \textbf{+4.76} & & \textbf{51.94} & \textbf{22.89} & \textbf{+14.85}\\
    \bottomrule
    \end{tabular}
    }
    \caption{Validation set performance taken across all tasks on NYUv2 and Cityscapes using CNN backbones. Values in bold indicate the best value in a given column for multitask models in SS and MS configurations. \s{IS IT possible to include Pascal as well? why do you only report with transformer for Pascal?}}
    \label{tab:sota_nyu}
\end{table}

\s{~\Cref{tab:sota_nyu} shows the results of CTAL$_{SS}$ and CTAL$_{MS}$ against all our baselines using CNN backbones. The table is divided into sections to separate the traditional STL and MTL baselines, the SS models, and the MS models. As we can see, for both datasets, we achieve considerably higher performance for all task metrics in SS and MS configurations. We can even see that our CTAL$_{SS}$ is competitive to MTI-Net in NYUv2 despite not having deep supervision on task predictions from multiple scales. On Cityscapes, we can also see that MTI-Net struggles in the simpler 2-task setting with smaller input image resolution. The results of MTI-Net have likely not been reported previously on Cityscapes as it is susceptible to overfitting. Despite our efforts to mitigate overfitting (i.e. spatial dropout, warm restart scheduler, data augmentation, architectural modifications, hyperparameter tuning), we were unable to achieve better performance than what is seen in ~\cref{tab:sota_nyu}. However, our CTAL$_{MS}$, which also employs deep supervision on multi-scale initial predictions, does not exhibit the same overfitting behaviour. In fact, we further improve performance over CTAL$_{SS}$.}

~\Cref{tab:sota_nyu} presents the results of CTAL$_{SS}$ and CTAL$_{MS}$ compared to all baselines using CNN backbones. The table is organized into sections for the traditional STL and MTL baselines, SS models, and MS models. As shown, our method achieves significantly higher performance across all task metrics in both SS and MS configurations for both datasets. Notably, CTAL$_{SS}$ remains competitive with MTI-Net on NYUv2, despite lacking deep supervision from multiple scales. On Cityscapes, MTI-Net underperforms in the simpler 2-task setting with a smaller input resolution, likely due to overfitting. Previous results for MTI-Net on Cityscapes have likely not been reported, possibly for this reason. Despite efforts to prevent overfitting—such as spatial dropout, warm restart scheduler, data augmentation, architectural modifications, and hyperparameter tuning—the performance of MTI-Net did not surpass what is shown in ~\cref{tab:sota_nyu}. However, CTAL$_{MS}$, which applies deep supervision on multi-scale initial predictions, avoids this overfitting and further improves performance compared to CTAL$_{SS}$.

% \usepackage{multirow}
\begin{table}[]
\centering
\setlength{\tabcolsep}{4pt}
\resizebox{0.4\textwidth}{!}{%
\begin{tabular}{ccccccc}
\toprule
& \multicolumn{6}{c}{\textit{NYUv2 (Transformer)}}\\
\cline{2-7}\\[-2ex]
\multirow{2}{*}{Model} & \multicolumn{1}{l}{Params} & \multicolumn{1}{l}{FLOPs} & \textbf{SemSeg} & \textbf{Depth} & \textbf{Normals} & \multirow{2}{*}{$\Delta_m \uparrow$} \\
%\cline{4-6}\\[-2ex]
 & (M) & (G) & mIoU $\uparrow$ & rErr $\downarrow$ & mErr $\downarrow$ &  \\
 \midrule
STL & 166.8 & 148.2 & 55.78 & 0.1570 & 24.59 & +0.00 \\
MTL & 68.8 & 60.5 & 57.07 & 0.1500 & 24.76 & +2.03 \\
\midrule
PAD-Net & 65.6 & 57.4 & \underline{57.40} & \textbf{0.1469} & \underline{23.70} & \underline{+4.32} \\
PAP-Net & \textbf{55.8} & \textbf{49.2} & 56.80 & 0.1500 & 23.73 & +3.26 \\
CTAL$_{SS}$ & \underline{57.0} & \underline{49.9} & \textbf{58.06} & \underline{0.1491} & \textbf{23.27} & \textbf{+4.83}\\
\bottomrule
\end{tabular}}
\caption{Validation set performance taken across all tasks on NYUv2 using transformer backbones. Values in bold and underline indicate the best and seconds best value respectively in a given column for multitask models.}
\label{tab:txf_nyu}
\end{table}

\begin{table}[]
\centering
\setlength{\tabcolsep}{2pt}
\resizebox{0.48\textwidth}{!}{%
\begin{tabular}{ccccccccc}
\toprule
& \multicolumn{8}{c}{\textit{PASCAL-Context (Transformer)}}\\
\cline{2-9}\\[-2ex]
\multirow{2}{*}{Model} & \multicolumn{1}{l}{Params} & \multicolumn{1}{l}{FLOPs} & \textbf{SemSeg} & \textbf{HPSeg} & \textbf{Sal} & \textbf{Normals} & \textbf{Edge} & \multirow{2}{*}{$\Delta_m \uparrow$}\\ 
%\cline{4-8}\\[-2ex]
 & (M) & (G) & mIoU $\uparrow$ & mIoU $\uparrow$ & maxF $\uparrow$ & mErr $\downarrow$ & Loss $\downarrow$ &  \\
 \midrule
STL & 278.0 & 477.5 & 74.07 & 64.28 & 83.95 & 16.03 & 0.0241 & +0.00 \\
MTL & 108.7 & 82.1 & 72.99 & 60.13 & 82.93 & 16.76 & 0.0236 & -2.40 \\
\midrule
PAD-Net & 112.3 & 88.5 & \underline{73.27} & 60.22 & \underline{83.40} & \underline{16.44} & \underline{0.0235} & -1.86 \\
PAP-Net & \textbf{100.5} & \underline{64.9} & 72.75 & \underline{60.32} & 83.22 & 16.52 & \underline{0.0235} & \underline{-1.83} \\
CTAL$_{SS}$ & \underline{114.1} & \textbf{62.2} & \textbf{73.98} & \textbf{61.14} & \textbf{83.50} & \textbf{15.96} & \textbf{0.0210} & \textbf{+1.46}\\
\bottomrule
\end{tabular}
}
\caption{Validation set performance taken across all tasks on PASCAL-Context using transformer backbones. Values in bold and underline indicate the best and seconds best value respectively in a given column for multitask models. \s{also highlight the best in the first two columns, Clarify why it is used only for Pascal.  CTAL (Ours)?? differne naming compare to other tables. I expect o see ss and ms. }}
\label{tab:txf_pascal}
\end{table}

\s{~\Cref{tab:txf_nyu} and~\cref{tab:txf_pascal} show the results of CTAL$_{SS}$ against our baselines using transformer backbones on the NYUv2 and PASCAL-Context datasets respectively. The results for NYUv2 seem consistent with the results obtained using CNN backbones, where the MTL models all perform well and CTAL performs the best. As we can see for PASCAL-Context, the STL baseline is much more competitive in this higher parameter and data regime. The comparative reduction in performance of the other MTL methods could be attributed to the dampening of the MTL regularization benefits seen in smaller data regimes. Additionally, the inclusion of more tasks can lead to increased task competition, which is also known as negative transfer~\cite{caruana1997multitask}. However, the results show that CTAL is resistant to these factors and is the only model that achieves a positive MTL gain while using significantly fewer model parameters and FLOPs than STL.}

~\Cref{tab:txf_nyu,tab:txf_pascal} present the results of CTAL$_{SS}$ compared to baselines using transformer backbones on the NYUv2 and PASCAL-Context datasets, respectively. The NYUv2 results align with those obtained using CNN backbones, where the MTL models perform well, and CTAL achieves the best results. On PASCAL-Context, however, the STL baseline is more competitive in this higher-parameter and larger-data regime. The reduced performance of other MTL methods may be due to the diminished regularization benefits of MTL in larger data settings. Additionally, the inclusion of more tasks can increase task competition, leading to negative transfer~\cite{caruana1997multitask}. Nevertheless, the results demonstrate that CTAL is resistant to these challenges, being the only model to achieve a positive MTL gain while using significantly fewer parameters and FLOPs than STL.

% Since this study targets light-weight CNN-based approaches, we do not compare with the transformer-based models in this parameter regime. However, we would like to note that when equipped with comparable Transformer backbones, the current CNN-based baselines perform very similarly to the InvPT and TaskPrompter~\cite{taskprompter2023} (see Table \ref{tab:comp} in the appendix). 

 \subsection{Ablation Study}
\begin{table}[htb!]
    \setlength{\tabcolsep}{4.0pt}
    \centering
    \resizebox{0.46\textwidth}{!}{%
    \begin{tabular}{cccccccccc}
    \toprule
     & & \multicolumn{4}{c}{\textit{NYUv2 (CNN)}} & & \multicolumn{3}{c}{\textit{Cityscapes (CNN)}}\\
     \cline{3-6}\cline{8-10}\\[-2ex]
      \multirow{2}{*}{Model} & & \textbf{SemSeg} & \textbf{Depth} & \textbf{Normals} & \multirow{2}{*}{$\Delta_m \uparrow$} & & \textbf{SemSeg} & \textbf{Depth} & \multirow{2}{*}{$\Delta_m \uparrow$}\\
       & & mIoU $\uparrow$ &  relErr $\downarrow$ & mErr $\downarrow$ & & & mIoU $\uparrow$ &  relErr $\downarrow$ &\\
     %\midrule
     \cline{1-1}\cline{3-6}\cline{8-10}\\[-2ex]
     STL & & 49.23 & 0.1636 & 23.15 & +0.00 & & 48.89 & 29.91 & \hspace{5pt}+0.00\\ 
     %\midrule
     \cline{1-1}\cline{3-6}\cline{8-10}\\[-2ex]
    Concat &  & 50.01 & 0.1634 & 23.65 & -0.15 & & 50.33 & 26.92 & \hspace{5pt}+6.47 \\ 
     CTAL$_{SS}$ & & 51.59 & 0.1607 & 22.84 & \textbf{+2.64} & & 51.36 & 23.84 & \textbf{+12.67} \\
     CTAL$_{MS}$ & & 52.70 & 0.1529 & 22.99 & \textbf{+4.76} & & 51.94 & 22.89 & \textbf{+14.85} \\
    \bottomrule
    \end{tabular}
    }
    \caption{Effectiveness of the different configurations of CTAL for both NYUv2 and Cityscapes datasets. The Concat model refers to the traditional task-prediction distillation framework that simply concatenates the features of the initial task prediction for cross-task distillation.}
    \label{tab:modular_ablation}
\end{table}

\s{To observe the effects of CTAL in isolation for our SS and MS configurations (i.e., CTAL$_{SS}$ and CTAL$_{MS}$), we follow the same procedure a~\cite{xu2018pad}, where we evaluate the performance of the base task-prediction distillation architecture without CTAL or the CSF blocks, and use a simple concatenation operation for the cross-task distillation step. This means that we simply take the features of each initial task prediction, concatenate them together, and pass that block to the final decoders. This model is named ``Concat" in~\cref{tab:modular_ablation}. Next, we evaluate the performance change from adding CTAL in the single-scale variation (CTAL$_{SS}$), and then with cross-scale fusion in the multi-scale variation (CTAL$_{MS}$). As we can see in Table~\ref{tab:modular_ablation}, the standard task-prediction distillation framework using a concatenation operation does not outperform the STL baseline for NYUv2 and achieves a +6.47\% MTL gain on Cityscapes. With CTAL$_{SS}$, we can see a +2.64\% and +12.67\% improvement in the MTL gain for each dataset compared to the STL baseline. Finally, having CTAL$_{MS}$ with deep supervision at multiple scales with CSF, we see further MTL gains with +4.76\% and +14.85\% on NYUv2 and Cityscapes respectively. Overall, we see large MTL gains across both datasets using our SS and MS configurations.}\s{WHy shown only for these two datasets? }

To isolate the effects of CTAL in our SS and MS configurations (i.e., CTAL$_{SS}$ and CTAL$_{MS}$), we follow the procedure from~\cite{xu2018pad}, where the base task-prediction distillation architecture is evaluated without CTAL or CSF blocks, using a simple concatenation operation for the cross-task distillation step. This ``Concat" model, shown in ~\cref{tab:modular_ablation}, concatenates the features of each initial task prediction and passes the combined block to the final decoders. Next, we evaluate the impact of adding CTAL in the single-scale configuration (CTAL$_{SS}$), followed by cross-scale fusion in the multi-scale configuration (CTAL$_{MS}$). As seen in ~\cref{tab:modular_ablation}, the standard task-prediction distillation framework with concatenation achieves respectable baseline results. With CTAL$_{SS}$, we realize improvements across all metrics for NYUv2 and Cityscapes. Finally, CTAL$_{MS}$ achieves further improvements in MTL gain for both datasets. Overall, substantial MTL gains are observed across both datasets using our SS and MS configurations.

\subsection{Resource Analysis}
\label{resource}

\begin{table}[htb!]
\setlength{\tabcolsep}{4pt}
\centering
\resizebox{0.35\textwidth}{!}{%
\begin{tabular}{cccccccc}
\toprule
\multirow{2}{*}{Model} & & \multirow{2}{*}{\textbf{Scale}} & \textbf{Rel} & \multirow{2}{*}{\textbf{FLOPs}} & \multirow{2}{*}{\textbf{Time (s)}} & \multirow{2}{*}{$\Delta_m \uparrow$}\\
& & & \textbf{Param.} & & & \\
\hline\\[-2ex]
MTL & & 1/4 & 0.346 & \hspace{5pt}42G & \hspace{5pt}2.82 & -1.89\\
PAP-Net & & 1/4 & 0.404 & 521G & 17.73 & +0.04 \\
PAD-Net & & 1/4 & 1.023 & 484G & \hspace{5pt}6.13 & +0.27 \\
%EMA-Net (C Att) & 0.497 & +1.17 & - \\
\cline{3-7}\\[-2ex]
\multirow{3}{*}{CTAL$_{SS}$} & & 1/4 & 0.513 & 537G & 13.85&\textbf{+2.64} \\
 & & 1/6 & 0.504 & 194G & \hspace{5pt}5.97 & \textbf{+1.70} \\
 & & 1/8 & 0.500 & 124G & \hspace{5pt}4.37 & \textbf{+1.50} \\
\midrule
MTI-Net & & 1/4 & 1.070 & \hspace{5pt}65G & \hspace{5pt}4.45 & +3.04 \\
\cline{3-7}\\[-2ex]
\multirow{3}{*}{CTAL$_{MS}$} & & 1/4 & 0.944 & 525G & 14.26 &\textbf{+4.76} \\
 & & 1/6 & 0.935 & 190G & \hspace{5pt}6.48 & \textbf{+3.62} \\
 & & 1/8 & 0.932 & 117G & \hspace{5pt}5.23 & \textbf{+3.43} \\
\bottomrule
\end{tabular}
}
\caption{Resource analysis of SS and MS models on NYUv2 using a CNN backbone. ``Scale" represents the feature scale (relative to the input image) used for task-prediction distillation. ``Rel Param.'' represents the number of parameters relative to the STL baseline. ``FLOPs" represents the number of floating-point operations used in a forward pass. ``Time" refers to the wall clock time required to process the entire validation set in seconds. Finally, ``$\Delta_m$" represents the MTL gain relative to the STL baseline.}
\label{tab:res}
\end{table}

\s{As mentioned earlier, the main issue with task-prediction distillation methods has been their relative resource requirements when using CNN backbones. Specifically, the number of FLOPs explodes when distilling features at larger scales; which is common among CNN feature extractors. Although there are many optimization opportunities to reduce the effect of additional FLOPs, like optimizing algorithmically~\cite{fawzi2022discovering}, improving hardware utilization~\cite{kljucaric2023deep}, and leveraging sparse matrix operations~\cite{gao2023systematic}, the most effective tactic is to manipulate feature scales. In~\cref{tab:param}, we demonstrate how we can manipulate the feature scales of the CNN backbone to realize massive reductions in model parameters and FLOPs while still outperforming all other baselines at larger scales. However, simply using a transformer backbone in a higher parameter regime permits the use of information rich features at low scales. This drastically reduces the number of FLOPs required for task-prediction distillation, while introducing a negligible number of additional model parameters, as seen in~\cref{tab:txf_nyu,tab:txf_pascal}. Despite the inability of the other task prediction distillation methods to obtain a positive MTL gain in~\cref{tab:txf_pascal}, they still outperform the MTL baseline while using comparable model parameters and FLOPs. This demonstrates that task-prediction distillation is a viable and lightweight technique when using transformer based backbones in multitask learning. Specifically for CTAL, the results demonstrate several efficiency benefits from our proposed method. This can be attributed to how we organize the features to allow for more efficient parameter usage. This also reduces the risk of overfitting, as seen in the results of the Cityscapes dataset. When equipped with a transformer backbone, our efficiency benefits are more pronounced, leading to much fewer FLOPs than the large STL and the lightweight MTL baselines.}

As mentioned earlier, a key challenge for task-prediction distillation methods has been their resource demands when using CNN backbones, particularly the explosion of FLOPs when distilling features at larger scales, which is common with CNN feature extractors. Although various optimization strategies can reduce additional FLOPs—such as algorithmic optimizations~\cite{fawzi2022discovering}, improved hardware utilization~\cite{kljucaric2023deep}, and sparse matrix operations~\cite{gao2023systematic}—the most effective approach is manipulating feature scales. In ~\cref{tab:res}, we show how adjusting the CNN backbone's feature scales significantly reduces model parameters and FLOPs while still outperforming all other baselines at larger scales. In contrast, transformer backbones in higher-parameter regimes allow for information-rich features at lower scales, drastically reducing FLOPs for task-prediction distillation and introducing minimal additional parameters, as seen in ~\cref{tab:txf_nyu,tab:txf_pascal}. Despite other task-prediction distillation methods failing to achieve a positive MTL gain in ~\cref{tab:txf_pascal}, they still outperform the MTL baseline with comparable parameters and FLOPs. This shows that task-prediction distillation is a viable, lightweight technique with transformer backbones in multitask learning. For CTAL specifically, the results highlight several efficiency advantages, stemming from our feature organization that enables more efficient parameter usage and reduces the risk of overfitting, as demonstrated in the Cityscapes dataset results. When paired with transformer backbones, these efficiency benefits become even more pronounced, leading to significantly fewer FLOPs compared to both large STL and lightweight MTL baselines.
% In general, as the trend of using more model capacity for better performance persists, we must be more conscious of how we can optimize efficiency so that these models can be deployed in real-world settings with much stricter memory limitations~\cite{menghani2023efficient}.

\subsection{Qualitative Analysis}
\begin{figure}
    \centering
    \includegraphics[width=0.98\linewidth]{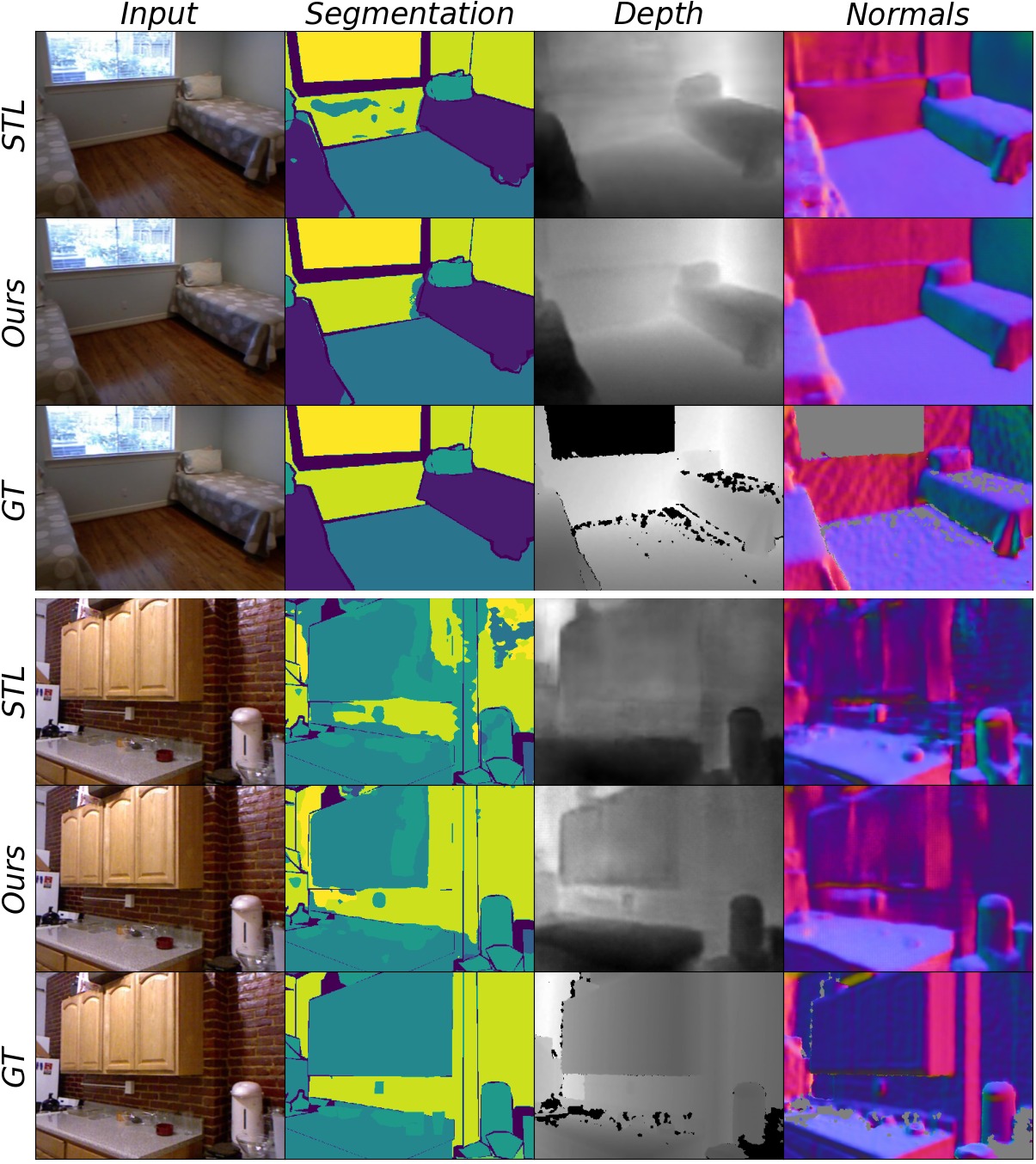}
    \caption{A visual comparison of the predictions from the single task baseline (STL) and CTAL$_{MS}$ (Ours)  \s{CHANGE it here and on the figure to CTAL$_{ms}$}. The two images and the ground truths (GT) are from the validation set of NYUv2.}
    \label{fig:qual}
\end{figure}

~\Cref{fig:qual} is a visualization of the predictions of our CTAL$_{MS}$ and the STL baseline on images from the validation set of NYUv2. Generally speaking, our model produces significantly fewer artifacts in the segmentation maps, and generates smoother depth and surface normal maps compared to STL. Specifically, we experience less warping of objects and we generalize better in undefined regions (i.e., windows).

% \subsection{Limitations}
% \label{limitations}
% Due to computational constraints, we were unable to experiment on datasets with more labeled examples, higher image resolutions, and/or more tasks, such as Taskonomy \cite{zamir2018taskonomy} or PASCAL-Context \cite{mottaghi_cvpr14}. However, our experiments are sufficiently diversified by scene type (indoor and outdoor), task types, and image resolution to demonstrate our model's ability to improve multitask learning in low-parameter regimes; while also remaining consistent with the experimental setups from other popular MTL works, like MTAN \cite{liu2019end}, PAD-Net \cite{kendall2018multi}, PAP-Net \cite{zhang2019pattern}. Evaluating CTAL's local, global, and cross-task pattern modelling capabilities in larger parameter regimes with more tasks would be an interesting future direction. It would also be interesting to test other techniques to mitigate the impact of FLOPs on wall-clock time, like sparsifying affinity matrices.

\section{Conclusion}
% also mention future steps for sparsifying affinity matrices to speed up computations
\s{We proposed our novel CTAL module for task prediction distillation. It is the first cross-task distillation module that can explicitly and exhaustively model all local and long-range feature-pair relationships intra- and inter-task. Impressively, we accomplish this using significantly fewer model parameters and FLOPs than single-task learning while achieving the best multitasking performance on complex indoor and outdoor scenes. Our experiments with a transformer backbone have demonstrated that task-prediction distillation is still a viable performance enhancer for dense prediction in MTL, while requiring similar or even fewer computational resources than the simplest hard-parameter sharing multitask architecture.}

We proposed the novel CTAL module for task prediction distillation, the first cross-task distillation module capable of explicitly and exhaustively modeling all local and long-range feature-pair relationships both intra- and inter-task. Remarkably, we achieve this with significantly fewer parameters and FLOPs than single-task learning, while delivering the best multitasking performance on complex indoor and outdoor scenes. Our experiments with a transformer backbone demonstrate that task-prediction distillation remains a strong performance enhancer for dense prediction in MTL, requiring similar or even fewer computational resources than the simplest hard-parameter sharing multitask architectures.

%%%%%%%%% REFERENCES
{\small
\bibliographystyle{ieee_fullname}
\bibliography{egbib}

\begin{thebibliography}{10}\itemsep=-1pt

\bibitem{bruggemann2021exploring}
David Br{\"u}ggemann, Menelaos Kanakis, Anton Obukhov, Stamatios Georgoulis, and Luc Van~Gool.
\newblock Exploring relational context for multi-task dense prediction.
\newblock In {\em Proceedings of the IEEE/CVF international conference on computer vision}, pages 15869--15878, 2021.

\bibitem{caruana1997multitask}
Rich Caruana.
\newblock Multitask learning.
\newblock {\em Machine learning}, 28(1):41--75, 1997.

\bibitem{chen2017deeplab}
Liang-Chieh Chen, George Papandreou, Iasonas Kokkinos, Kevin Murphy, and Alan~L Yuille.
\newblock Deeplab: Semantic image segmentation with deep convolutional nets, atrous convolution, and fully connected crfs.
\newblock {\em IEEE transactions on pattern analysis and machine intelligence}, 40(4):834--848, 2017.

\bibitem{chen2018gradnorm}
Zhao Chen, Vijay Badrinarayanan, Chen-Yu Lee, and Andrew Rabinovich.
\newblock Gradnorm: Gradient normalization for adaptive loss balancing in deep multitask networks.
\newblock In {\em International conference on machine learning}, pages 794--803. PMLR, 2018.

\bibitem{chollet2017xception}
Fran{\c{c}}ois Chollet.
\newblock Xception: Deep learning with depthwise separable convolutions.
\newblock In {\em Proceedings of the IEEE conference on computer vision and pattern recognition}, pages 1251--1258, 2017.

\bibitem{Cordts2016Cityscapes}
Marius Cordts, Mohamed Omran, Sebastian Ramos, Timo Rehfeld, Markus Enzweiler, Rodrigo Benenson, Uwe Franke, Stefan Roth, and Bernt Schiele.
\newblock The cityscapes dataset for semantic urban scene understanding.
\newblock In {\em Proc. of the IEEE Conference on Computer Vision and Pattern Recognition (CVPR)}, 2016.

\bibitem{dosovitskiy2020image}
Alexey Dosovitskiy, Lucas Beyer, Alexander Kolesnikov, Dirk Weissenborn, Xiaohua Zhai, Thomas Unterthiner, Mostafa Dehghani, Matthias Minderer, Georg Heigold, Sylvain Gelly, et~al.
\newblock An image is worth 16x16 words: Transformers for image recognition at scale.
\newblock {\em arXiv preprint arXiv:2010.11929}, 2020.

\bibitem{eigen2015predicting}
David Eigen and Rob Fergus.
\newblock Predicting depth, surface normals and semantic labels with a common multi-scale convolutional architecture.
\newblock In {\em Proceedings of the IEEE international conference on computer vision}, pages 2650--2658, 2015.

\bibitem{pascal-voc-2010}
M. Everingham, L. Van~Gool, C.~K.~I. Williams, J. Winn, and A. Zisserman.
\newblock The {PASCAL} {V}isual {O}bject {C}lasses {C}hallenge 2010 {(VOC2010)} {R}esults.
\newblock http://www.pascal-network.org/challenges/VOC/voc2010/workshop/index.html.

\bibitem{fawzi2022discovering}
Alhussein Fawzi, Matej Balog, Aja Huang, Thomas Hubert, Bernardino Romera-Paredes, Mohammadamin Barekatain, Alexander Novikov, Francisco~J R~Ruiz, Julian Schrittwieser, Grzegorz Swirszcz, et~al.
\newblock Discovering faster matrix multiplication algorithms with reinforcement learning.
\newblock {\em Nature}, 610(7930):47--53, 2022.

\bibitem{fu2019dual}
Jun Fu, Jing Liu, Haijie Tian, Yong Li, Yongjun Bao, Zhiwei Fang, and Hanqing Lu.
\newblock Dual attention network for scene segmentation.
\newblock In {\em Proceedings of the IEEE/CVF conference on computer vision and pattern recognition}, pages 3146--3154, 2019.

\bibitem{gao2023systematic}
Jianhua Gao, Weixing Ji, Fangli Chang, Shiyu Han, Bingxin Wei, Zeming Liu, and Yizhuo Wang.
\newblock A systematic survey of general sparse matrix-matrix multiplication.
\newblock {\em ACM Computing Surveys}, 55(12):1--36, 2023.

\bibitem{gao2019nddr}
Yuan Gao, Jiayi Ma, Mingbo Zhao, Wei Liu, and Alan~L Yuille.
\newblock Nddr-cnn: Layerwise feature fusing in multi-task cnns by neural discriminative dimensionality reduction.
\newblock In {\em Proceedings of the IEEE/CVF conference on computer vision and pattern recognition}, pages 3205--3214, 2019.

\bibitem{he2016deep}
Kaiming He, Xiangyu Zhang, Shaoqing Ren, and Jian Sun.
\newblock Deep residual learning for image recognition.
\newblock In {\em Proceedings of the IEEE conference on computer vision and pattern recognition}, pages 770--778, 2016.

\bibitem{kingma2014adam}
Diederik~P Kingma and Jimmy Ba.
\newblock Adam: A method for stochastic optimization.
\newblock {\em arXiv preprint arXiv:1412.6980}, 2014.

\bibitem{kljucaric2023deep}
Luke Kljucaric and Alan~D George.
\newblock Deep learning inferencing with high-performance hardware accelerators.
\newblock {\em ACM Transactions on Intelligent Systems and Technology}, 14(4):1--25, 2023.

\bibitem{liu2022auto}
Shikun Liu, Stephen James, Andrew~J Davison, and Edward Johns.
\newblock Auto-lambda: Disentangling dynamic task relationships.
\newblock {\em arXiv preprint arXiv:2202.03091}, 2022.

\bibitem{liu2019end}
Shikun Liu, Edward Johns, and Andrew~J Davison.
\newblock End-to-end multi-task learning with attention.
\newblock In {\em Proceedings of the IEEE/CVF conference on computer vision and pattern recognition}, pages 1871--1880, 2019.

\bibitem{liu2022swin}
Ze Liu, Han Hu, Yutong Lin, Zhuliang Yao, Zhenda Xie, Yixuan Wei, Jia Ning, Yue Cao, Zheng Zhang, Li Dong, et~al.
\newblock Swin transformer v2: Scaling up capacity and resolution.
\newblock In {\em Proceedings of the IEEE/CVF conference on computer vision and pattern recognition}, pages 12009--12019, 2022.

\bibitem{loshchilov2016sgdr}
Ilya Loshchilov and Frank Hutter.
\newblock Sgdr: Stochastic gradient descent with warm restarts.
\newblock {\em arXiv preprint arXiv:1608.03983}, 2016.

\bibitem{maninis2019attentive}
Kevis-Kokitsi Maninis, Ilija Radosavovic, and Iasonas Kokkinos.
\newblock Attentive single-tasking of multiple tasks.
\newblock In {\em Proceedings of the IEEE/CVF conference on computer vision and pattern recognition}, pages 1851--1860, 2019.

\bibitem{misra2016cross}
Ishan Misra, Abhinav Shrivastava, Abhinav Gupta, and Martial Hebert.
\newblock Cross-stitch networks for multi-task learning.
\newblock In {\em Proceedings of the IEEE conference on computer vision and pattern recognition}, pages 3994--4003, 2016.

\bibitem{mottaghi_cvpr14}
Roozbeh Mottaghi, Xianjie Chen, Xiaobai Liu, Nam-Gyu Cho, Seong-Whan Lee, Sanja Fidler, Raquel Urtasun, and Alan Yuille.
\newblock The role of context for object detection and semantic segmentation in the wild.
\newblock In {\em IEEE Conference on Computer Vision and Pattern Recognition (CVPR)}, 2014.

\bibitem{Silberman:ECCV12}
Pushmeet~Kohli Nathan~Silberman, Derek~Hoiem and Rob Fergus.
\newblock Indoor segmentation and support inference from rgbd images.
\newblock In {\em ECCV}, 2012.

\bibitem{sandler2018mobilenetv2}
Mark Sandler, Andrew Howard, Menglong Zhu, Andrey Zhmoginov, and Liang-Chieh Chen.
\newblock Mobilenetv2: Inverted residuals and linear bottlenecks.
\newblock In {\em Proceedings of the IEEE conference on computer vision and pattern recognition}, pages 4510--4520, 2018.

\bibitem{sinodinos2022attentive}
Dimitrios Sinodinos and Narges Armanfard.
\newblock Attentive task interaction network for multi-task learning.
\newblock In {\em 2022 26th International Conference on Pattern Recognition (ICPR)}, pages 2885--2891. IEEE, 2022.

\bibitem{9336293}
Simon Vandenhende, Stamatios Georgoulis, Wouter Van~Gansbeke, Marc Proesmans, Dengxin Dai, and Luc Van~Gool.
\newblock Multi-task learning for dense prediction tasks: A survey.
\newblock {\em IEEE Transactions on Pattern Analysis and Machine Intelligence}, 44(7):3614--3633, 2022.

\bibitem{vandenhende2020mti}
Simon Vandenhende, Stamatios Georgoulis, and Luc Van~Gool.
\newblock Mti-net: Multi-scale task interaction networks for multi-task learning.
\newblock In {\em Computer Vision--ECCV 2020: 16th European Conference, Glasgow, UK, August 23--28, 2020, Proceedings, Part IV 16}, pages 527--543. Springer, 2020.

\bibitem{vaswani2017attention}
Ashish Vaswani, Noam Shazeer, Niki Parmar, Jakob Uszkoreit, Llion Jones, Aidan~N Gomez, {\L}ukasz Kaiser, and Illia Polosukhin.
\newblock Attention is all you need.
\newblock {\em Advances in neural information processing systems}, 30, 2017.

\bibitem{wang2020deep}
Jingdong Wang, Ke Sun, Tianheng Cheng, Borui Jiang, Chaorui Deng, Yang Zhao, Dong Liu, Yadong Mu, Mingkui Tan, Xinggang Wang, et~al.
\newblock Deep high-resolution representation learning for visual recognition.
\newblock {\em IEEE transactions on pattern analysis and machine intelligence}, 43(10):3349--3364, 2020.

\bibitem{woo2018cbam}
Sanghyun Woo, Jongchan Park, Joon-Young Lee, and In~So Kweon.
\newblock Cbam: Convolutional block attention module.
\newblock In {\em Proceedings of the European conference on computer vision (ECCV)}, pages 3--19, 2018.

\bibitem{xie2017aggregated}
Saining Xie, Ross Girshick, Piotr Doll{\'a}r, Zhuowen Tu, and Kaiming He.
\newblock Aggregated residual transformations for deep neural networks.
\newblock In {\em Proceedings of the IEEE conference on computer vision and pattern recognition}, pages 1492--1500, 2017.

\bibitem{xin2022current}
Derrick Xin, Behrooz Ghorbani, Justin Gilmer, Ankush Garg, and Orhan Firat.
\newblock Do current multi-task optimization methods in deep learning even help?
\newblock {\em Advances in Neural Information Processing Systems}, 35:13597--13609, 2022.

\bibitem{xu2018pad}
Dan Xu, Wanli Ouyang, Xiaogang Wang, and Nicu Sebe.
\newblock Pad-net: Multi-tasks guided prediction-and-distillation network for simultaneous depth estimation and scene parsing.
\newblock In {\em Proceedings of the IEEE Conference on Computer Vision and Pattern Recognition}, pages 675--684, 2018.

\bibitem{invpt2022}
Hanrong Ye and Dan Xu.
\newblock Inverted pyramid multi-task transformer for dense scene understanding.
\newblock In {\em ECCV}, 2022.

\bibitem{taskprompter2023}
Hanrong Ye and Dan Xu.
\newblock Taskprompter: Spatial-channel multi-task prompting for dense scene understanding.
\newblock In {\em ICLR}, 2023.

\bibitem{zhang2019pattern}
Zhenyu Zhang, Zhen Cui, Chunyan Xu, Yan Yan, Nicu Sebe, and Jian Yang.
\newblock Pattern-affinitive propagation across depth, surface normal and semantic segmentation.
\newblock In {\em Proceedings of the IEEE/CVF conference on computer vision and pattern recognition}, pages 4106--4115, 2019.

\end{thebibliography}
}

\clearpage
\appendix
\onecolumn

\section*{Appendix}
\section{Dataset Details}
\textbf{NYUv2} contains 1449 densely labelled RGB-depth images of indoor scenes. The raw dataset contains images with incomplete depth values; which are masked during training. The tasks associated with this dataset are 13-label semantic segmentation, depth estimation, and surface normals prediction. The dataset does not contain surface normal labels out-of-the-box, so following the literature~\cite{liu2019end}, we used the pseudo ground surface normals data obtained from~\cite{eigen2015predicting}, which include some incomplete values at the same locations as the corresponding depth maps. The training and validation sets contain 795 and 654 images respectively, and the resolution of the images is $288\times 384$.

\textbf{Cityscapes} is a larger dataset containing 3475 outdoor urban street scenes with fine annotations taken from 50 cities over several months of the year. From the set of fine annotations, we have 2975 train and 500 validation images. The tasks associated with this dataset are 19-label semantic segmentation and depth estimation. The labels used are from their official documentation that group several labels into a void class, and specify 19 other labels that should be used during training. The resolution of the images is $128\times 256$.

\textbf{PASCAL-Context}~\cite{mottaghi_cvpr14} is an even larger dataset derived from the PASCAL VOC 2010 challenge \cite{pascal-voc-2010}, containing pixel-wise annotations for 10,103 images. These images cover a wide range of indoor and outdoor scenes with various objects. The dataset includes 4,998 training and 5,105 validation images. The tasks associated this dataset are 21-label semantic segmentation, human parts segmentation, edge detection, saliency, and surface normals. The resolution of the images varies, so they are padded and scaled to $512\times 512$.

\section{Related Works} 
% Introduce and briefly explain some of the popular ones, like PAD-Net, MTI-Net, etc
PAD-Net~\cite{xu2018pad} is the first work to popularize the ``task prediction distillation" framework. Their cross-task distillation module uses EM attention, which can capture local patterns intra- and inter-task, but lacks the ability to model long-range dependencies. PAP-Net~\cite{zhang2019pattern} is another distillation algorithm that explicitly models feature similarities, known as ``task affinity" using MM attention. Although they capture local and long-range dependencies intra-task, their simplistic cross-task diffusion mechanism inhibits inter-task pattern propagation. MTI-Net~\cite{vandenhende2020mti} extends distillation to multiple feature scales, which is known as ``multi-scale task-prediction distillation". The cross-task distillation algorithm they use for each scale is the same one used by PAD-Net (i.e., EM attention). The number of additional model parameters for generating initial task predictions and cross-task distillation modules at multiple scales makes this method very inefficient as the input image size and number of tasks increase. Also, this framework isn't suited for most ViT-based backbones that output features as a single scale.

The aforementioned cross-task distillation algorithms are inspired by the attention mechanism~\cite{vaswani2017attention}; which allows networks to place greater emphasis on certain parts of an input that are important for the downstream task. For dense vision tasks, it has been shown that attending to features in the spatial and/or channel dimensions leads to significant performance improvements~\cite{woo2018cbam,fu2019dual}. Consequently, these notions have been extended to the MTL domain, which explored different ways of modelling cross-task patterns using attention~\cite{liu2019end,xu2018pad,zhang2019pattern,vandenhende2020mti,sinodinos2022attentive}.

Other recent MTL works for dense scene predictions include 
ATRC~\cite{bruggemann2021exploring}, InvPt~\cite{invpt2022}, and TaskPrompter~\cite{taskprompter2023}. ATRC applies a neural architecture search (NAS) to learn a branching structure that considers the global features, local features, source label, and target labels between every possible combination of task pairs. Although this study provides interesting insights into optimal task interactions, it is difficult to justify its use in a real-world setting because it takes an incredible amount of resources to train, and scales very poorly with more tasks. Hanrong Ye and Dan Xu~\cite{invpt2022,taskprompter2023} create their own multitask network based on the Vision Transformer (ViT)~\cite{dosovitskiy2020image}. The added model capacity allows them to explicitly model local and global relationships between tasks. Despite both being encoder-focused works, they compare their results to the decoder-focused distillation algorithms using CNN backbones. Although they perform an unfair comparison, the broader consideration is that encoder- and decoder-focused algorithms are not mutually exclusive and can be used in a complimentary fashion. Additionally, these encoder-focused methods are not practical for real-world application because they require a handcrafted design for a given backbone; which are constantly evolving. Decoder-focused methods, like task-prediction distillation methods, are modular and can be used with an arbitrary pretrained backbone.

\section{Tasks and Performance Metrics}
\textbf{Semantic Segmentation} refers to the task of assigning a class label to each pixel in an image. During training, the objective is to minimize the depth-wise cross-entropy loss between the predicted labels $\hat{y}$, and the targets $y$, for all $N$ pixels:

\begin{equation}
    \mathcal{L}_{Semantic} = -\frac{1}{N}\sum_{n \epsilon N} y_n \text{log}(\hat{y}_n)
\label{sem_loss}
\end{equation}

We also evaluate our models on mean intersection over union (mIoU) and absolute pixel accuracy. Given the true positives (TP), false positives (FP), and false negatives (FN) for each image, we compute mIoU as follows:

\begin{equation}
    mIoU = \frac{1}{N}\sum_{n \in N}\frac{TP_n}{TP_n + FP_n + FN_n}
\label{iou}
\end{equation}

\textbf{Human Parts Segmentation} is defined and evaluated in the exact same way as the semantic segmentation task. The only difference between these tasks is the nature of the assigned labels. For human parts segmentation, pixels are assigned a label based on a human body part rather than labels of objects (i.e., car, road, building).

\textbf{Depth Estimation} involves predicting the depth values at each pixel. During training, we aim to minimize the absolute error ($L_1$ norm) of the predicted values $\hat{d}$, and the targets $d$:

\begin{equation}
    \mathcal{L}_{Depth} = \sum_{n \epsilon N}||d_n-\hat{d_n}||
\label{abs_err}
\end{equation}

We also report on the relative depth error:

\begin{equation}
    Error_{rel} = \sum_{n \epsilon N}\frac{||d_n-\hat{d}_n||}{d_n}
\label{rel_err}
\end{equation}

\textbf{Surface Normals} prediction involves estimating the direction perpendicular to the surface of objects in an image; making it useful for acquiring geometric and structural scene information. We train the model to minimize the element-wise dot product between the normalized predictions $\hat{s}$, and the targets $s$:

\begin{equation}
    \mathcal{L}_{Normals} = -\frac{1}{N}\sum_{n \epsilon N}s_n\cdot\hat{s}_n
\label{norm_loss}
\end{equation}

For evaluating surface normals prediction performance, we also consider the mean angular distance between $\hat{s}$ and $s$. Angular distance is the arccosine of the sum of the element-wise product of $\hat{s}$ and $s$, as seen in Equation~\ref{angle_distance}. We also report the proportion of predictions that fall within 11.25, 22.5, and 30.0 degrees of error.

\begin{equation}
    D_\theta = \text{arccos}(\sum_{n \epsilon N}\hat{s}_n\cdot s_n)
\label{angle_distance}
\end{equation}
\\
% \begin{equation}
%     \mathcal{L}_{Semantic} = -\frac{1}{N}\sum_{n \epsilon N} y_n \text{log}(\hat{y}_n)
% \label{sem_loss}
% \end{equation}
% We also evaluate our models on mean intersection over union (mIoU) and absolute pixel accuracy. However, mIoU is a much better indicator of semantic understanding. Given the true positives (TP), false positives (FP), and false negatives (FN) for each image, we compute mIoU as follows:
%
% \begin{equation}
%     mIoU = \frac{1}{N}\sum_{n \in N}\frac{TP_n}{TP_n + FP_n + FN_n}
% \label{iou}
% \end{equation}
\textbf{Saliency detection} involves identifying the most visually important regions in an image. The model is trained to predict a saliency map $\hat{S}$, highlighting areas that are likely to attract human attention. During training, we minimize the pixel-wise binary cross-entropy loss between the predicted saliency map $\hat{S}$ and the ground truth saliency map $S$. We evaluate using the max F-measure, which evaluates the balance between precision and recall across different thresholds applied to the predicted saliency map. The formula for the F-measure is:

\begin{equation}
    F_\beta = \frac{(1 + \beta^2) \cdot \text{Precision} \cdot \text{Recall}}{(\beta^2 \cdot \text{Precision}) + \text{Recall}}
\end{equation}

For the max-F measure, you compute the F-measure across multiple thresholds ($\tau$) and take the maximum:

\begin{equation}
    \text{max-}F_\beta = \max \left( F_\beta(\tau_1), F_\beta(\tau_2), \dots, F_\beta(\tau_n) \right)
\end{equation}

\textbf{Edge detection} involves detecting boundaries between different regions in an image. The model is trained to predict binary edge maps $\hat{E}$, where pixels corresponding to edges are labeled as 1 and others as 0. We minimize the binary cross-entropy loss during training and evaluate using the validation loss.
%
% \begin{equation}
%     \mathcal{L}_{Normals} = -\frac{1}{N}\sum_{n \epsilon N}s_n\cdot\hat{s}_n
% \label{norm_loss}
% \end{equation}
%
% For evaluating surface normals prediction performance, we also consider the mean angular distance between $\hat{s}$ and $s$. Angular distance is the arccosine of the sum of the element-wise product of $\hat{s}$ and $s$, as seen in Equation~\ref{angle_distance}. We also report the proportion of predictions that fall within 11.25, 22.5, and 30.0 degrees of error.
%
% \begin{equation}
%     D_\theta = \text{arccos}(\sum_{n \epsilon N}\hat{s}_n\cdot s_n)
% \label{angle_distance}
% \end{equation}
\\
\textbf{MTL Gain}~\cite{maninis2019attentive} is an aggregate measure of the overall multitask improvement of method $m$ with respect to a single task learning baseline $b$ for all tasks $t \in [1,N]$, as seen in Equation \ref{deltam}.
\begin{equation}
    \Delta_m = \frac{1}{N}\sum_t^N(-1)^{l_t}(M_{m,t}-M_{b,t})/M_{b,t}
\label{deltam}
\end{equation}
where $l_t = 1$ if a lower value of metric $M$ is favorable, and 0 otherwise. We will treat $\Delta_m$ as a percentage in our evaluation. Although we use multiple metrics per task throughout our evaluations, we want to make sure that every task is weighed evenly when calculating $\Delta_m$ by selecting a single metric per task that best demonstrates generalization performance. Consequently, to compute $\Delta_m$, we will use mIoU for segmentation, relative error for depth, mean angular distance for surface normals, max F-measure for saliency, and validation loss for edge detection. We also show that we still achieve superior MTL gain using other combinations of metrics in the~\cref{tab:all_city} and~\cref{tab:all_nyu}. In our results, the metrics where larger values are favourable are denoted with ($\uparrow$) and smaller values with ($\downarrow$).

\section{Results with Additional Metrics}
In ~\cref{tab:all_city} and ~\cref{tab:all_nyu}, we can see that in addition to the results in main paper, we also outperform all other models using other metrics for the Cityscapes and NYUv2 datasets. Therefore, using any combination of evaluation metrics to compute the multitask gain ($\Delta_m$) will show we still achieve the best overall multitask performance.

\begin{table}[htb!]
    \centering
    \setlength{\tabcolsep}{4pt}
    \resizebox{0.7\textwidth}{!}{%
    \begin{tabular}{c c c c c c c c c c c c c c c} %{\textwidth}
    \toprule
      & & \multicolumn{11}{c}{\textit{NYUv2 (CNN)}}\\
      \cline{2-13}\\[-2ex]
      & \multicolumn{2}{c}{\textbf{SemSeg}} & & \multicolumn{2}{c}{\textbf{Depth}} & & \multicolumn{4}{c}{\textbf{Normals}} & & \multirow{2}{*}{$\Delta_m \uparrow$}\\ 
     \cline{2-3}\cline{5-6}\cline{8-11}\\[-2ex]
      % & & & & & & & Angle Distance & \multicolumn{3}{c}{Within t\degree} \\ 
      % Model & \multicolumn{2}{c}{(Higher Better)} & & \multicolumn{2}{c}{(Lower Better)} & & (Lower Better) & \multicolumn{3}{c}{(Higher Better)} \\ 
      Model & mIoU $\uparrow$ & pixAcc $\uparrow$ & & relErr $\downarrow$ & mErr $\downarrow$& & mErr $\downarrow$& 11.25 $\uparrow$& 22.5 $\uparrow$& 30 $\uparrow$ & &\\
     \midrule
     STL & 49.23 & 72.83 & & 0.1636 & 0.3853 & & 23.15 & 35.18 & 62.50 & 73.48 & & +0.00\\ 
     MTL  & 49.25 & 72.90 & & 0.1658 & 0.3896 & & 24.16 & 30.80 & 57.92 & 70.41 & & -1.89\\ 
      \midrule
     PAD-Net & 50.23 & 73.46 & & 0.1622 & 0.3814 & & 23.63 & 32.44 & 59.51 & 71.68 & & +0.27\\
     PAP-Net & 50.00 & 73.25 & & 0.1615 & 0.3876 & & 23.78 & 31.90 & 58.89 & 71.22 & & +0.04\\
     %EMA-Net (C Att) & 49.40 & 73.07 & & 0.1609 & 0.3864 & & 23.48 & 33.34 & 69.71 & 71.72 & & +1.17\\ 
     CTAL$_{SS}$ & \textbf{51.59} & \textbf{74.14} & & \textbf{0.1607} & \textbf{0.3808} & & \textbf{22.84} & \textbf{35.14} & \textbf{62.06}& \textbf{73.40} & & \textbf{+2.64}\\
     \midrule
      MTI-Net & 51.51 & 74.50 & & 0.1538 & 0.3650 & & 23.50 & 34.16 & 60.85 & 72.31 & & +3.04\\
     CTAL$_{MS}$ & \textbf{52.70} & \textbf{75.09} & & \textbf{0.1529} & \textbf{0.3630} & & \textbf{22.99} & \textbf{35.59} & \textbf{62.25}& \textbf{73.28} & & \textbf{+4.76}\\
    \bottomrule
    \end{tabular}
    }
    \caption{Validation set performance taken across all tasks on NYUv2 and Cityscapes using CNN backbones. Values in bold indicate the best value in a given column for multitask models in SS and MS configurations.}
    \label{tab:all_nyu}
\end{table}

\begin{table*}[htb!]
    \centering
    \setlength{\tabcolsep}{4pt}
    \begin{tabular}{c c c c c c c c c c c c c c} %{\textwidth}
    \toprule
      & \multicolumn{2}{c}{\textbf{Sem. Seg.}} & & \multicolumn{2}{c}{\textbf{Depth}}\\ 
     \cline{2-3}
     \cline{5-6}\\[-2ex]
      Model & mIoU $\uparrow$ & pixAcc $\uparrow$ & & relErr $\downarrow$ & mErr $\downarrow$\\
     \midrule
     STL & 48.89 & 90.87 & & 29.91 & 1.296\\ 
     MTL  & 49.78 & 91.07 & & 31.80 & 1.155\\ 
      \midrule
     PAP-Net & 50.82 & 91.19 & & 26.97 & 1.135\\
     PAD-Net & 50.67 & 91.24 & & 27.37 & 1.136\\
     %EMA-Net (C Att) & 49.40 & 73.07 & & 0.1609 & 0.3864 & & 23.48 & 33.34 & 69.71 & 71.72 & & +1.17\\ 
     CTAL$_{SS}$ & \textbf{51.36} & \textbf{91.34} & & \textbf{23.84} & \textbf{1.119}\\
     \midrule
      MTI-Net & 51.77 & 91.13 & & 29.90 & 1.141\\
     CTAL$_{MS}$ & \textbf{51.94} & \textbf{91.27} & & \textbf{22.89} & \textbf{1.127}\\
    \bottomrule
    \end{tabular}
    \caption{Validation set performance taken across all tasks on NYUv2. Values in bold indicate the best value in a given column for multitask models in SS and MS configurations.}
    \label{tab:all_city}
\end{table*}

\section{Implementation Details}
\label{implementation}
All CNN models are equipped with a pre-trained HRNet18~\cite{wang2020deep} multiscale feature extractor backbone. The single-scale variants will use a fused version of the input features following the aforementioned CSF procedure. All transformer models are equipped with a pre-trained SwinV2-s~\cite{liu2022swin} backbone. Since the output of the transformer backbone is an aggregated feature representation, only SS models are evaluated. The output heads for the initial predictions include two residual blocks~\cite{he2016deep} followed by an output convolution layer. The initial predictions used for task-prediction distillation are the outputs of the second residual block. The final output heads for the CNN models use the same architecture as the heads used for the initial predictions, but for Transformer models, we use a DeepLab~\cite{chen2017deeplab} head to get the final predictions since it is still a popular choice for dense prediction tasks like segmentation. The implementation code for all baseline networks is taken from~\cite{vandenhende2020mti}, except for PAP-Net which we carefully implemented ourselves since there was not a publicly available implementation.

\section{Hyperparameters}

We train our models using an Adam~\cite{kingma2014adam} optimizer with a weight decay of $1\times 10^{-4}$. The learning rates are $1\times 10^{-4}$, $5\times 10^{-4}$, and $2\times 10^{-5}$ for NYUv2, Cityscapes, and PASCAL-Context respectively. We performed a small learning rate search (within the range of 1e-2 and 1e-5) for each model to ensure that this configuration was favourable for all baselines. We also use a cosine annealing learning rate scheduler~\cite{loshchilov2016sgdr} for smooth convergence. Multi-scale models tend to converge early for Cityscapes, so for them, we used a cosine annealing learning rate scheduler with warm restarts~\cite{loshchilov2016sgdr} to promote exploration and escape local minima. For all datasets, we use a batch size of 8, a blending factor $\gamma=0.05$ (like PAP-Net) and filter size $f=3$ for all our models. The values for $\gamma$ and $f$ were not tuned for each dataset, and our models show little sensitivity to these parameters. We train for 200, 75, and 70 epochs on NYUv2, Cityscapes, and PASCAL-Context respectively using a single NVIDIA RTX A5000 GPU. The training time per run in this setup was approximately 4 hours for NYUv2, 1 hour for Cityscapes, and 9 hours for PASCAL-Context.

\section{Hyperparameter Sensitivity}
\subsection{Blending Factor $\gamma$}
\begin{table}[htb!]
    \centering
    \setlength{\tabcolsep}{4pt}
    \begin{tabular}{cccccc}
    \toprule
      \multirow{2}{*}{Model} & \multirow{2}{*}{$\gamma$} & \textbf{Sem. Seg.} & \textbf{Depth} & \textbf{Normals} & \textbf{MTL Gain}\\
       & & mIoU $\uparrow$ &  relErr $\downarrow$ & mErr $\downarrow$ & $\Delta_m \uparrow$\\
     \midrule
     \multirow{3}{*}{EMA-Net (SS)} & 0.025 & 50.68 & 0.1608 & 22.53 & +2.45\\
      & 0.050 & 51.59 & 0.1607 & 22.84 & +2.64 \\
     & 0.100 & 52.49 & 0.1631 & 22.84 & +2.76\\
     \midrule
     \multirow{3}{*}{EMA-Net (MS)} & 0.025 & 51.71 & 0.1526 & 23.01 & +4.12 \\
      & 0.050 & 52.70 & 0.1529 & 22.99 & +4.76 \\
      & 0.100 & 53.44 & 0.1557 & 23.06 & +4.59\\
    \bottomrule
    \end{tabular}
    \caption{Validation set performance taken across all tasks on NYUv2 for different values of blending factor $\gamma$.}
    \label{tab:g}
\end{table}
As we can see from Table~\ref{tab:g}, there is noticeable variability in segmentation performance when using different blending factors ($\gamma$) for both SS and MS models. However, we can see that the performance of the other tasks compensates accordingly, such that the overall MTL gain does not change significantly. This is consistent with the expected competitive nature between tasks when training multitask systems.

\subsection{Filter Size $f$}
\begin{table}[htb!]
    \centering
    \setlength{\tabcolsep}{4pt}
    \begin{tabular}{cccccc}
    \toprule
      \multirow{2}{*}{Model} & \multirow{2}{*}{$f$} & \textbf{Sem. Seg.} & \textbf{Depth} & \textbf{Normals} & \textbf{MTL Gain}\\
       & & mIoU $\uparrow$ &  relErr $\downarrow$ & mErr $\downarrow$ & $\Delta_m \uparrow$\\
     \midrule
     \multirow{3}{*}{CTAL$_{SS}$} & 3 & 51.59 & 0.1607 & 22.84 & +2.64 \\
      & 5 & 52.19 & 0.1630 & 22.89 & +2.50\\
     & 7 & 51.58 & 0.1640 & 22.79 & +2.03\\
     \midrule
     \multirow{3}{*}{CTAL$_{MS}$} & 3 & 52.70 & 0.1529 & 22.99 & +4.76 \\
      & 5 & 53.21 & 0.1547 & 23.06 & +4.64\\
      & 7 & 52.70 & 0.1557& 23.14 & +3.97\\
    \bottomrule
    \end{tabular}
    \caption{Validation set performance taken across all tasks on NYUv2 for different filter sizes $f$.}
    \label{tab:f}
\end{table}
In Table~\ref{tab:f}, we can see that using different filter sizes ($f$) for cross-task pattern modelling, we do not see a significant drop in performance between $f=3$ and $f=5$. However, using too large of a filter size, i.e., $f=7$, we can expect a drop in performance.

\section{Results With Standard Deviation}
\label{sigma}
Tables \ref{tab:nyu_sigma} and \ref{tab:city_sigma} contain identical results presented in the main paper, but with the addition of the standard deviation across all runs. This is to provide a notion of statistical confidence for our results. The formula used to compute the standard deviation is as follows:

\begin{equation}
    \sigma = \sqrt{\frac{1}{N-1} \sum_{i=1}^{N} (x_i - \bar{x})^2}
\end{equation}

\begin{table}[htb!]
    \setlength{\tabcolsep}{4.0pt}
    \centering
 %   \resizebox{0.48\textwidth}{!}{%
    \begin{tabular}{cccccc}
    \toprule
     & & \multicolumn{4}{c}{\textit{NYUv2}}\\
     \cline{3-6}\\[-2ex]
      \multirow{2}{*}{Model} & & \textbf{Sem. Seg.} & \textbf{Depth} & \textbf{Normals} & \multirow{2}{*}{$\Delta_m \uparrow$}\\
       & & mIoU ($\sigma$) $\uparrow$ &  relErr ($\sigma$) $\downarrow$ & mErr ($\sigma$) $\downarrow$ & \\
     %\midrule
     \cline{1-1}\cline{3-6}\\[-2ex]
     STL & & 49.23 (0.29) & 0.1636 (0.0024) & 23.15 (0.09) & +0.00\\
     MTL & & 49.25 (0.43) & 0.1658 (0.0028) & 24.16 (0.05) & -1.89\\
     %\midrule
     \cline{1-1}\cline{3-6}\\[-2ex]
     PAP-Net & & 50.00 (0.49) & 0.1615 (0.0043) & 23.78 (0.07) & +0.04\\
     PAD-Net & & 50.23 (0.41) & 0.1622 (0.0016) & 23.63 (0.06) & +0.27\\
     CTAL$_{SS}$ & & \textbf{51.59} (0.33) & \textbf{0.1607} (0.0008) & \textbf{22.84} (0.06) & \textbf{+2.64}\\
     \cline{1-1}\cline{3-6}\\[-2ex]
     MTI-Net & & 51.51 (0.63) & 0.1538 (0.0011) & 23.50 (0.04) & +2.64\\
     CTAL$_{MS}$ & & \textbf{52.70} (0.34) & \textbf{0.1529} (0.0027) & \textbf{22.99} (0.06) & \textbf{+4.76}\\
    \bottomrule
    \end{tabular}
%    }
   \caption{Average validation set performance taken across all tasks on NYUv2 for 3 runs. Values in bold indicate the best value in a given column for multitask models in SS and MS configurations. Values in brackets indicate the standard deviation across three runs.}
    \label{tab:nyu_sigma}
\end{table}

\begin{table}[htb!]
    \setlength{\tabcolsep}{4.0pt}
    \centering
    % \resizebox{0.4\textwidth}{!}{%
    \begin{tabular}{ccccc}
    \toprule
     & & \multicolumn{3}{c}{\textit{Cityscapes}}\\
     \cline{3-5}\\[-2ex]
      \multirow{2}{*}{Model} & & \textbf{Sem. Seg.} & \textbf{Depth} & \multirow{2}{*}{$\Delta_m \uparrow$}\\
       & & mIoU ($\sigma$) $\uparrow$ &  relErr ($\sigma$) $\downarrow$ &\\
     %\midrule
     \cline{1-1}\cline{3-5}\\[-2ex]
     STL & & 48.89 (0.74) & 29.91 (0.88) & \hspace{5pt}+0.00\\
     MTL & & 49.78 (0.36) & 31.80 (0.48) & \hspace{5pt}-2.25\\
     %\midrule
     \cline{1-1}\cline{3-5}\\[-2ex]
     PAP-Net & & 50.82 (0.72) & 26.97 (0.67) & \hspace{5pt}+6.89\\
     PAD-Net & & 50.67 (0.44) & 27.37 (0.52) & \hspace{5pt}+6.07\\
     CTAL$_{SS}$ & & \textbf{51.36} (0.64) & \textbf{23.84} (0.58) & \textbf{+12.67} \\
     \cline{1-1}\cline{3-5}\\[-2ex]
     MTI-Net & & 51.77 (0.84) & 29.90 (0.48) & +2.96 \\
     CTAL$_{MS}$ & & \textbf{51.94} (0.26) & \textbf{22.89} (0.48) & \textbf{+14.85} \\
    \bottomrule
    \end{tabular}
    % }
    \caption{Average validation set performance taken across all tasks on Cityscapes for 3 runs. Values in bold indicate the best value in a given column for multitask models in SS and MS configurations. Values in brackets indicate the standard deviation across three runs.}
    \label{tab:city_sigma}
\end{table}
\end{document}